%% file: main.tex
\documentclass{article}
\usepackage{iclr2026_conference,times}
\usepackage[T1]{fontenc}
\input{math_commands.tex}

\usepackage{hyperref}
\usepackage{url}
\usepackage{graphicx}
\usepackage{float}

\usepackage{booktabs}
\usepackage{amsmath,amssymb}
\usepackage[font=small,labelfont=bf,skip=4pt]{caption}
\usepackage{xcolor}
\usepackage{colortbl}
\newcommand{\exq}[1]{\cellcolor{gray!15}#1}
\usepackage{tabularx}
\usepackage[most]{tcolorbox}
\usepackage{seqsplit}

\definecolor{genframe}{RGB}{60,105,155}
\definecolor{genback}{RGB}{234,241,249}
\newtcolorbox{genseq}[1]{
breakable, enhanced, colback=genback, colframe=genframe,
coltitle=white, fonttitle=\bfseries\footnotesize, fontupper=\ttfamily\scriptsize,
boxrule=0.6pt, arc=1.5pt, left=5pt, right=5pt, top=3pt, bottom=3pt, boxsep=2pt,
title={#1}}

\title{Generative modeling of intrinsically disordered protein regions by reinforcing sparse autoencoder features}

\author{
\textbf{Jason X. Liu}\textsuperscript{1},
\textbf{Sebastian Ibarraran}\textsuperscript{1},
\textbf{Frank Hu}\textsuperscript{1},
\textbf{Soojung Yang}\textsuperscript{1}, 
\textbf{Xinyu A. Feng}\textsuperscript{1}, \\
\textbf{Abigail Park}\textsuperscript{1}, 
\textbf{Anagha Aneesh}\textsuperscript{1},
\textbf{Lacramioara Bintu}\textsuperscript{1}, 
\textbf{Alexander R. Dunn}\textsuperscript{1$\ast$}, \\
\textbf{Grant M. Rotskoff}\textsuperscript{1$\ast$} \\
\textsuperscript{1}Stanford University \\
\texttt{\{jxliu2, ardunn, rotskoff\}@stanford.edu}
}

\iclrfinalcopy %
\begin{document}
\maketitle

\lhead{ } %

\renewcommand{\thefootnote}{}
\footnotetext{$^\ast$Corresponding authors} %
\footnotetext{ }
\renewcommand{\thefootnote}{\arabic{footnote}}

\begin{abstract}

Intrinsically disordered protein regions (IDRs) play central roles in cellular processes such as transcriptional regulation, signal transduction, and subcellular localization, yet their functional design remains challenging. Structure-based design methods do not readily apply to IDRs, and existing protein language models are trained on full-length protein sequences, thus learning a prior that is biased towards folded domains. Here, we present \textbf{IDiom}, an autoregressive protein language model trained on IDiom-DB, a dataset of 54 million predicted IDRs curated from the AlphaFold Database. IDiom generates diverse sequences that recapitulate the composition, patterning, motifs, and predicted disorder of natural IDRs. To control function-associated sequence patterns, we also introduce \textbf{reinforcement learning with sparse autoencoder features (RL-SAE)}, a post-training method that rewards the generation of sequences that activate specified feature sets. Across eight IDR design tasks, RL-SAE sequences activate, on average, 90\% of 30 targeted features, compared to 24\% for activation steering. We demonstrate that RL-SAE improves the predicted subcellular localization and transcriptional activity of generated IDRs compared to steering and supervised fine-tuning, and enables features associated with distinct biological functions to be combined within individual sequences. Thus, IDiom and RL-SAE enable interpretable and composable IDR design through explicit control of function-associated sequence features. More broadly, RL-SAE could extend to other protein design settings where interpretable features provide useful design targets. Code is available at \url{https://github.com/rotskoff-group/idiom}.

\end{abstract}

\section{Introduction}

Intrinsically disordered protein regions (IDRs) are ubiquitous, constituting roughly 30\% of residues in the human proteome, with over 60\% of human proteins containing at least one IDR \citep{tesei_conformational_2024, alderson_systematic_2023}. It is now clear that IDRs play essential roles in almost every aspect of cellular function. Because IDRs do not adopt well-defined folds, these sequences can act as flexible linkers \citep{gonzalez-foutel_conformational_2022}, multivalent signaling hubs \citep{wright_intrinsically_2015}, and drivers of biomolecular condensate formation \citep{wei_phase_2017}, and they play crucial roles in diverse biological processes such as signal transduction, transcriptional regulation, chromatin organization, and subcellular localization \citep{holehouse_molecular_2024}.

Rational design of IDRs would unlock new opportunities in bioengineering, including tunable condensate formation, precise regulation of cell signaling, and targeted protein localization \citep{tesei_computational_2026}. Tremendous progress has been seen recently in machine learning-based design of folded proteins, driven by accurate structure prediction \citep{jumper_highly_2021, abramson_accurate_2024} and structure-based generative models \citep{ahern_atom-level_2026, geffner_-proteina_2026}. However, direct application of these approaches to IDRs remains challenging due to their structurally disordered nature.

Protein language models (PLMs) have recently been explored for applications in protein design. These models learn rich evolutionary statistics over amino acid sequences by training on large corpora of protein sequences from databases such as UniProt \citep{the_uniprot_consortium_uniprot_2025}, enabling them to generate novel proteins and functional variants \citep{hayes_simulating_2025, bhatnagar_scaling_2025}. However, these models are primarily trained on full-length proteins and possess a generative prior that is biased towards folded domains \citep{ferruz_protgpt2_2022}. Training on an IDR-specific corpus would instead focus the prior on the sequence statistics of disordered regions, specializing the generative model for IDR design.

Control over the incorporation of interpretable sequence features into generated proteins also remains a key challenge. PLMs can be fine-tuned for functional design using online reinforcement learning (RL) against computational oracles \citep{stocco_guiding_2025, angermueller_model-based_2020} and offline alignment with experimental datasets \citep{chennakesavalu_aligning_2025, ibarraran_efficient_2026}. However, these methods do not explicitly identify or control the functional sequence-level patterns that are present in the generated proteins. Sparse autoencoders (SAEs) decompose PLM representations into features associated with interpretable sequence patterns \citep{candido_language_2026, simon_interplm_2025}. The ability to target selected SAE features would enable explicit control over these patterns, allowing those associated with distinct biological functions to be composed.

\paragraph{Key contributions:}
Here, we address these gaps with three primary contributions:

\begin{itemize}
\item \textbf{IDiom, a protein language model for intrinsically disordered regions}, pretrained on IDiom-DB, a corpus of 54M predicted IDRs curated from the AlphaFold Database. IDiom enables the generation of diverse sequences that recapitulate natural IDR composition, patterning, motifs, and predicted disorder. 

\item \textbf{Reinforcement learning with sparse autoencoder features (RL-SAE)}, a post-training method that rewards the generation of sequences that activate specified sets of interpretable SAE features. Across eight design tasks, RL-SAE-generated sequences activate, on average, 90\% of 30 targeted features, compared to 24\% for activation steering.

\item \textbf{Computational experiments demonstrating selective and composable IDR design.} We show that RL-SAE improves the predicted subcellular localization and transcriptional activity of generated IDRs compared to activation steering and supervised fine-tuning, while enabling selective incorporation and combination of function-associated sequence features.

\end{itemize}

\section{Related work}

\paragraph{Design of intrinsically disordered regions}
Current sequence-based IDR design techniques primarily employ residue sampling strategies to construct disordered sequences and variants that satisfy imposed constraints such as compositional rules or sequence-level metrics \citep{pesce_design_2024, hunter_rational_2026}. However, these techniques impose disorder through hand-specified constraints and do not necessarily capture the evolutionary statistics of natural IDRs. In contrast, IDiom directly learns these sequence statistics, and with RL-SAE, this enables biological design using function-associated features rather than manually specified sequence rules.

\paragraph{Interpreting and steering protein language models}
Sparse autoencoders trained on PLM activations have been shown to find approximately monosemantic features corresponding to interpretable biological concepts \citep{candido_language_2026, simon_interplm_2025}, and activation steering along such feature directions has been applied to sequence design \citep{corominas_sparse_2025, villegasgarcia_interpreting_2025}. 
However, combining multiple steering directions does not ensure that generated sequences activate all targeted features. In contrast, RL-SAE activates a substantially greater fraction of target features than activation steering, 
and we show that this enables the design of IDRs that combine interpretable features associated with distinct biological functions, such as localization and transcriptional regulation. %

\paragraph{Reinforcement learning with sparse autoencoders}
Several recent works in the large language model literature have combined SAEs with reinforcement learning. For example, \cite{cho_control_2026} train a controller to select and amplify SAE feature directions during inference, and \cite{zhang_interpretable_2025} construct a reward model using SAE feature activations fit to human preference pairs for improved reward interpretability. Unlike these methods, RL-SAE directly rewards the activation of prespecified sets of target SAE features, allowing for explicit control over the incorporation of these features into generated sequences.

\begin{figure}[h!]\centering
\includegraphics[width=\textwidth]{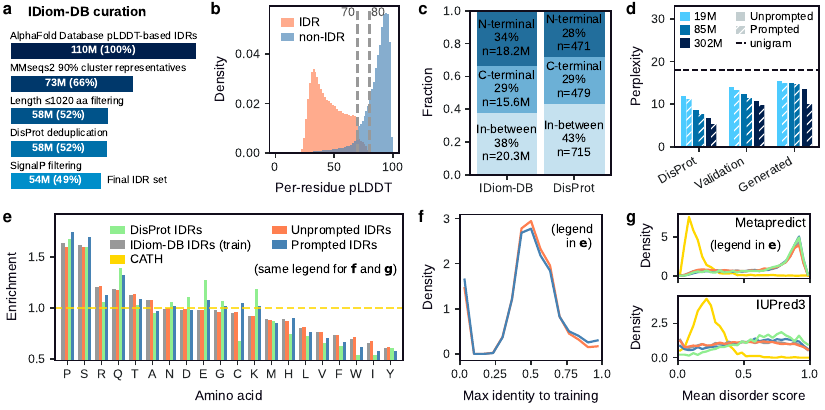}
\caption{
\textbf{Training on 54M curated IDRs enables the generation of diverse and biologically plausible IDRs.}
\textbf{(a)} IDiom-DB curation workflow. 
\textbf{(b)} AlphaFold2 pLDDT distribution of curated IDRs and non-IDRs. 
\textbf{(c)} Fraction of curated training IDRs and experimental DisProt IDRs which are located at the N-terminus, C-terminus, and internally. 
\textbf{(d)} IDiom model perplexity on DisProt, validation, and generated IDRs. Solid and hatched bars indicate unprompted and prompted IDRs, respectively. 
\textbf{(e)} Amino acid compositional enrichment for training, generated, and DisProt IDRs, relative to the folded CATH sequences' compositions. This panel's legend also applies to panels (f) and (g). 
\textbf{(f)} Sequence identity of generated IDRs relative to training set IDRs, calculated with MMseqs2. 
\textbf{(g)} Predicted disorder scores of training, generated, and DisProt IDRs, alongside CATH sequences (higher is more disordered). 
}
\label{fig:model}
\end{figure}

\section{Generative modeling of intrinsically disordered regions}
\label{sec:idiom}

\paragraph{A corpus of intrinsically disordered region sequences}

To curate the IDiom-DB dataset of IDR sequences for training, we use AlphaFold2 (AF2) predicted structures from the AlphaFold Database (AFDB) \citep{varadi_alphafold_2024}. We use low AF2 predicted local distance difference test (pLDDT) values as a predictor of disorder, as this has been demonstrated to correlate strongly with experimental measurements of disorder \citep{wilson_alphafold2_2022, zhao_comparative_2023}. We first cluster AFDB sequences at 90\% sequence identity using MMseqs2 \citep{steinegger_mmseqs2_2017} and then following \cite{tesei_conformational_2024}, we apply a windowed threshold at pLDDT = 70 to identify IDRs (Figure \ref{fig:model}b), with proteins containing multiple IDRs contributing multiple records. We discard IDRs shorter than 30 residues, IDRs residing in proteins longer than 1,020 residues, proteins that are entirely low-pLDDT, and predicted N-terminal signal peptides. This yields 54M IDRs and their positions within their full-length host proteins (Figure \ref{fig:model}a). Additional details are in Appendix~\ref{app:corpus}.

\paragraph{Architecture and training}
We pretrain IDiom, a decoder-only transformer, at three sizes: 19M, 85M, and 302M parameters (final model perplexities in Figure \ref{fig:model}d). We use a context length of 1,024 tokens, where each amino acid residue is assigned an individual token. Training uses a next-token objective on a fill-in-the-middle (FIM) transformed \citep{bavarian_efficient_2022} version of IDiom-DB, which places N- and C-terminal contexts before the IDR to enable generation conditioned on fixed flanking sequences. During training, we provide 50\% of sequences with their flanking contexts and 50\% without, enabling both context-prompted and unprompted IDR generation. All analysis in this paper focuses on the 302M parameter model (IDiom-300M). Additional details are in Appendix~\ref{app:model}.

\paragraph{IDiom generates diverse disordered regions}
To characterize the pretrained model, we generate 100,000 unprompted IDRs and 166,300 context-prompted IDRs, using the surrounding flanking contexts of 1,663 experimentally validated IDRs from DisProt as prompts (100 IDRs per prompt) \citep{nugnes_disprot_2026}. The distribution of maximum sequence identities between generated and training IDRs peaks broadly around 50\%, indicating that most generated sequences are substantially dissimilar to any sequence seen during training (Figure \ref{fig:model}f).

To assess the predicted disorder of the generated sequences, 
we compare them to the 1,663 experimentally validated IDRs from DisProt as well as 1,000 folded CATH S60 domains \citep{waman_cath_2025}. 
We find that across AF2-predicted pLDDTs \citep{mirdita_colabfold_2022} and the sequence-based disorder predictors Metapredict V3 \citep{lotthammer_metapredict_2024} and IUPred3 \citep{erdos_iupred3_2021}, generated sequences closely track the DisProt set, attaining substantially higher disorder scores than CATH domains (Figure~\ref{fig:model}g). 

We additionally compare IDiom-generated IDRs to DisProt and CATH sequences along compositional and patterning metrics. We find that IDiom IDRs are substantially enriched in the disorder-promoting proline and serine residues while being depleted in order-promoting aliphatics and aromatics \citep{theillet_alphabet_2013} (Figure \ref{fig:model}e). IDiom IDRs also recapitulate known patterning properties that govern IDR behavior, such as charge segregation \citep{das_conformations_2013}, reduced hydrophobic clustering \citep{zheng_hydropathy_2020}, and low sequence complexity \citep{wootton_statistics_1993}. Close agreement with DisProt is observed across all of these metrics. Additional details are in Appendix~\ref{app:generation}.

\section{Reinforcement learning with sparse autoencoder features}
\label{sec:rlsae}

\paragraph{Motivation}
The design of functional IDRs requires control over combinations of sequence-level features. SAEs provide interpretable features that can be used as design targets \citep{adams_mechanistic_2025, simon_interplm_2025}, but reliably inducing the activation of many features simultaneously within generated sequences remains challenging \citep{arad_saes_2025, xie_comparative_2025}. Reinforcement learning with sparse autoencoder features (RL-SAE) addresses this challenge by rewarding the model for generating sequences that activate a specified set of SAE features. Our experiments demonstrate that RL-SAE enables IDiom to generate sequences that activate tens of function-associated SAE features simultaneously and outperforms activation steering and supervised fine-tuning on biological design tasks.

\paragraph{A sparse dictionary over IDiom activations}
\label{sec:sae}
To perform RL-SAE on IDiom, we train IDiomSAE, a Top-$k$ SAE with $k = 32$ and 16,384 latents in the feature activation vector $\vz$  \citep{gao_scaling_2024}. IDiomSAE is trained on IDiom's residual stream activations at layer-18 of 24 and only uses activations from IDR residue positions, thus focusing its learned dictionary on sequence features within disordered regions. Given a residual stream activation $\vx$, IDiomSAE calculates:
\begin{equation}
\vz = \mathrm{TopK}\!\left(\mathrm{ReLU}\!\left(\mW_{\mathrm{enc}}(\vx - \vb_{\mathrm{dec}}) + \vb_{\mathrm{enc}}\right)\right), \qquad
\hat{\vx} = \mW_{\mathrm{dec}}\vz + \vb_{\mathrm{dec}},
\end{equation}
where the encoder and decoder weights and biases $\mW_{\mathrm{enc}}$, $\mW_{\mathrm{dec}}$, $\vb_{\mathrm{enc}}$, and $\vb_{\mathrm{dec}}$ are trained to minimize the fraction of variance unexplained, $\|\vx - \hat{\vx}\|_2^2 / \|\vx - \bar{\vx}\|_2^2$, 
where $\bar{\vx}$ is the mean activation over the training set. Additional details are in Appendix~\ref{app:sae}.

\paragraph{Identification of enriched SAE feature sets}
\label{sec:signatures}
Our approach is motivated by the idea that SAE features enriched in sequences with common labeled functionality capture the sequence patterns that are relevant to that functionality, and we use these features as reward targets in RL-SAE. 
To identify salient SAE features, we rank each feature $j$ by the log odds ratio (LOR) of its firing prevalence in the positively labeled set $\mathcal{P}$ to its prevalence within a negative set $\mathcal{N}$ \citep{candido_language_2026}. We count a feature as firing in a sequence $y$ if it activates at any residue position $t$: 
\begin{equation}
\mathrm{fire}_j(y) = \1\!\left[\max_t \vz_j(y)_t > 0\right], \qquad
\mathrm{LOR}_j = \log_2 \frac{\Pr_{\mathcal{P}}[\mathrm{fire}_j]\,/\,(1-\Pr_{\mathcal{P}}[\mathrm{fire}_j])}{\Pr_{\mathcal{N}}[\mathrm{fire}_j]\,/\,(1-\Pr_{\mathcal{N}}[\mathrm{fire}_j])}.
\label{eq:lor}
\end{equation}
We then rank features by LOR and choose the top-$N$ as the target set $S$ for RL-SAE. Additional details are in Appendix~\ref{app:signatures}.

\paragraph{A sparse feature-matching reward}
We reward generated completions $y$ for the fraction of target features that fire, with quadratic penalties on relative deviations from the target sequence length $L_{\mathrm{target}}$ and entropy $H_{\mathrm{target}}$. Thus, our reward $R(y)$ is: 
\begin{equation}
R(y) = R_S(y) - \left[\left(L(y) - L_{\mathrm{target}}\right)/L_{\mathrm{target}}\right]^2 - \left[\left(H(y) - H_{\mathrm{target}}\right)/H_{\mathrm{target}}\right]^2,
\label{eq:reward}
\end{equation}
where
\begin{equation}
R_S(y) = \frac{1}{N}\sum_{j \,\in\, S} \mathrm{fire}_j(y)
\end{equation}
is the fraction of top-$N$ features in $S$ that fire anywhere within $y$, $L(y)$ is the generated sequence length, and $H(y)$ is the Shannon entropy of the sequence's amino acid composition. Feature scoring is performed with a frozen IDiomSAE and uses activations from the frozen pretrained IDiom model.

\paragraph{Reinforcement learning optimization}
We post-train IDiom with group-relative policy optimization \citep{shao_deepseekmath_2024} to optimize $R(y)$ with KL regularization relative to the base model:
\begin{equation}
\mathcal{J}(\theta) = \E_{y \sim \pi_\theta}\!\left[R(y)\right] - \beta\, \KL\!\left(\pi_\theta \,\|\, \pi_{\mathrm{base}}\right)
\end{equation}
where $\pi_\theta$ is the trained policy, $\pi_\mathrm{base}$ is the base IDiom model, and $\beta$ controls the KL regularization strength.

\section{Experiments}

\subsection{Experimental setup}
\label{sec:experiments}

\paragraph{Functional sequence sets and oracles}
To demonstrate RL-SAE for IDR design, we focus on two IDR-driven biological functions: association with biomolecular compartments and transcriptional regulation. We use six human sequence sets with annotated localization to the nucleolus (629 sequences), chromosomes (846), nuclear speckles (586), stress granules (664), P-bodies (329), and the nuclear pore complex (101) \citep{kilgore_protein_2025}, with 100,000 randomly selected, length-matched IDRs from IDiom-DB as the negative set. For transcriptional regulation, we use experimentally validated human activation domain (AD, 288) and repressor domain (RD, 408) sequences from a large-scale gene regulatory activity tiling scan \citep{delrosso_large-scale_2023}, with 4,920 inert tiles from the same assay as the negative set.

We characterize the generated sequences using ProtGPS \citep{kilgore_protein_2025} for compartment association, DeepLoc 2.1 for subcellular localization \citep{odum_deeploc_2024}, PADDLE for transcriptional activation \citep{sanborn_simple_2021}, and known corepressor-recruiting motifs from \cite{delrosso_large-scale_2023} for repression. We also assess compartment-associated sequence motifs and we predict disorder using Metapredict V3 and IUPred3.
Dataset details are in Appendix~\ref{app:labels}.

\paragraph{RL-SAE training}
For all 8 sequence sets, we run RL-SAE and activation steering on the top-$N=30$ enriched features, referred to as top-30 sets. We also evaluate feature sets consisting of the top 30 enriched features that are unique to each compartment, referred to as private-30 sets. All post-training runs begin with the base IDiom model and generate unprompted IDRs. We target a sequence length of 100 residues for compartment-associated IDRs and 80 residues for effector domains to match the experimental tile length. We set the entropy target to 3.65 bits, the mean sequence entropy of IDRs across IDiom-DB. Training details and other top-$N$ values are in Appendix~\ref{app:rl}.

\paragraph{Baseline methods}
As a direct baseline, we compare RL-SAE to activation steering using the same target feature sets. We sum and normalize the IDiomSAE decoder vectors into the steering direction, which we add to the layer-18 residual stream at every token position, scaled by $\alpha$ times the original activation norm before renormalizing to that norm. We sweep $\alpha$ from 0 to 1.5, discarding cases whose generated sequences have mean entropy below the fifth percentile of IDiom-DB (3.08 bits), and report the remaining $\alpha$ which achieves the highest score (Appendix~\ref{app:baselines}). 

For the supervised fine-tuning (SFT) baseline, we fine-tune the base IDiom model using the positive IDR sequence sets from which the top-$N$ features were extracted. We additionally evaluate RL using ProtGPS scores as the reward for compartment association (RL-GPS). Additional details for baseline methods are in Appendix~\ref{app:baselines}.

\subsection{Results}

\subsubsection{Enriched features recover canonical IDR sequence grammars}
\label{sec:res_interp}

\begin{figure}[h!]\centering
\includegraphics[width=\textwidth]{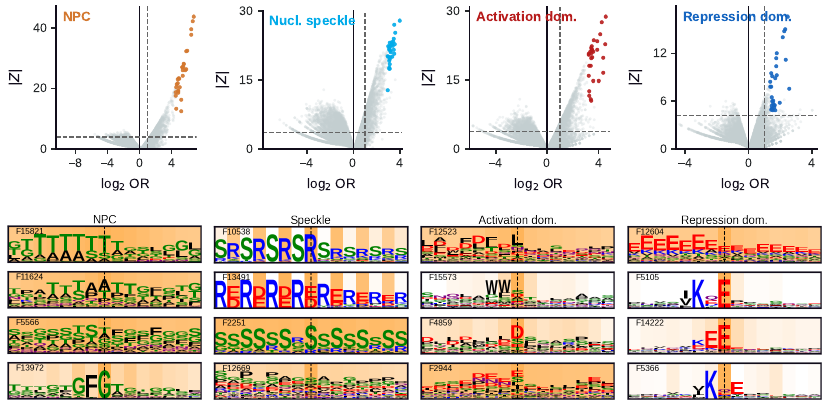}
\caption{
\textbf{Enriched SAE features recover canonical IDR sequence grammars.}
\textbf{Top row:} Statistical significance $|Z|$ versus log$_2$ odds ratio (LOR) for all 16,384 IDiomSAE features in nuclear pore complex, nuclear speckle, transcriptional activation, and repression sequence sets. Dashed lines mark LOR $=1$ (vertical) and the false discovery rate threshold of $10^{-3}$ (horizontal). Features in the upper right quadrant pass both thresholds. 
\textbf{Bottom row:} Information content logos for each set's top-four enriched features, showing residue likelihoods within a $\pm7$-residue window around peak firing positions (vertical dashed line position). Orange shading indicates the feature activation strength at each residue position. Logo construction is described in Appendix~\ref{app:galleries}.
}
\label{fig:signatures}
\end{figure}

\paragraph{Feature enrichment and sequence grammars}
Figure~\ref{fig:signatures} shows feature enrichment plots and sequence logos of the top-four enriched features for four example sequence sets.
We find that many of the top enriched features correspond to canonical IDR sequence grammars, such as FG/GLFG repeats in NPC sequences \citep{frey_fgfxfg_2009}, transportin-associated RS repeats of speckle sequences \citep{lai_transportin-sr2_2001}, the acidic exposure model patterning of activation domains \citep{staller_directed_2022}, and SUMOylation motifs in repression domains \citep{delrosso_large-scale_2023}. Additional enrichment and logo plots are in Appendix~\ref{app:galleries}. These examples show that feature enrichment identifies recognizable sequence grammars associated with the annotated functions, thus providing interpretable targets for RL-SAE. We next test whether RL-SAE increases target feature activation and improves predicted biological functionality in generated IDRs.

\subsubsection{Designing biomolecular condensate-associated IDRs}
\label{sec:res_ontarget}

\paragraph{Feature installation and compartment association}
First, we confirm the ability of RL-SAE to install the target feature set. We find that across the 8 sets, sequences generated after RL-SAE activate on average 90\% of the targeted top-30 features, compared to 24\% for activation steering. SFT activates only 15\% of the top-30 on average, and the natural sequences activate 21\%. Additional details are in Appendices~\ref{app:coverage} and~\ref{app:control}. 

Next, we assess whether the activation of targeted features is accompanied by known functional motifs and changes in predicted biological function. Table~\ref{tab:localization} shows the predicted ProtGPS localization scores for sequences generated by the different techniques, with predicted disorder for all methods given in Appendix~\ref{app:disorder}. We find that for all compartments except P-bodies, RL-SAE using the top-30 enriched features produces sequences with a higher mean predicted localization score than SFT, activation steering, and the natural IDR sequences. Composition-matched scrambles of the RL-SAE sequences show that these predicted scores are largely driven by specific grammars in all compartments except stress granules and P-bodies, whose predicted localization is partly compositional (Appendix~\ref{app:scramble}).

\begin{table}[h]\centering
\footnotesize
\setlength{\tabcolsep}{2.2pt}
\begin{tabular}{lcccccc}
\toprule
 & & & & & \multicolumn{2}{c}{RL-SAE} \\
\cmidrule(lr){6-7}
\textbf{Compartment} & Base & Nat. & SFT & Steering ($\alpha$) & top-30 & private-30 \\
\midrule
Nucleolus & $0.14_{.01}$ & $\underline{0.48}_{.02}$ & $0.30_{.01}$ & $0.41_{.01}$ (0.35) & $\mathbf{0.99}_{.00}$ & $0.94_{.07}$ \\
Chromosome & $0.26_{.01}$ & $\underline{0.52}_{.01}$ & $0.37_{.00}$ & $0.51_{.02}$ (0.21) & $\mathbf{0.71}_{.03}$ & $0.67_{.08}$ \\
Speckle & $0.00_{.00}$ & $\underline{0.15}_{.01}$ & $0.07_{.00}$ & $0.11_{.01}$ (0.20) & $\mathbf{0.36}_{.15}$ & $0.30_{.24}$ \\
\midrule
Stress gran. & $0.05_{.00}$ & $0.15_{.01}$ & $0.06_{.01}$ & $\underline{0.16}_{.01}$ (0.15) & $0.32_{.09}$ & $\mathbf{0.51}_{.36}$ \\
P-body & $0.17_{.01}$ & $\underline{0.27}_{.02}$ & $0.20_{.01}$ & $0.19_{.01}$ (0.00) & $0.05_{.02}$ & $\mathbf{0.32}_{.10}$ \\
\midrule
NPC & $0.08_{.01}$ & $\underline{0.42}_{.05}$ & $0.40_{.00}$ & $0.33_{.01}$ (0.20) & $0.84_{.11}$ & $\mathbf{0.90}_{.05}$ \\
\bottomrule
\end{tabular}
\caption{
\textbf{RL-SAE improves predicted association with target biomolecular compartments.}
ProtGPS-predicted biomolecular compartment association scores (0 to 1). Bold indicates the best score per row, underline indicates the best non-RL-SAE method. The private-30 column targets the top 30 private features of each compartment. Subscripted values are standard deviations over three seeds, except for natural IDRs (Nat.), which report the standard error of the mean (Appendix~\ref{app:stats}).
}
\label{tab:localization}
\end{table}

The experiments demonstrate that activation steering is limited by incomplete feature installation and decreased sequence entropy at higher steering strengths. While ProtGPS scores increase with $\alpha$ at low $\alpha$, the effect is limited prior to substantial sequence entropy collapse (Appendix~\ref{app:coverage}).
We find that SFT raises the score above the base model, but SFT scores remain below those of the natural sequences for all six compartments, potentially due to the limited dataset size.

Since the IDR sequences used for compartment association feature enrichment are extracted from the same full-length sequences used to train ProtGPS, there is circularity in using ProtGPS as a downstream evaluator. Therefore, we additionally score the generated sequences with DeepLoc 2.1, an orthogonal localization predictor \citep{odum_deeploc_2024}. Since DeepLoc only predicts coarse localization, we assess sequences on their predicted nuclear versus cytoplasmic localization. We find that RL-SAE targeting the top-30 or private-30 achieves the best or second-best score for the biologically expected label in all six compartments (Appendix~\ref{app:deeploc}). 
For P-body associating sequences, targeting the private-30 rather than the top-30 features increases the cytoplasmic DeepLoc prediction, consistent with the increased ProtGPS-predicted P-body score in Table~\ref{tab:localization}. 

In addition to ProtGPS and DeepLoc, we evaluate the generated sequences for the presence of several critical biological motifs: classical nuclear localization signals (cNLSs) for the nucleolus and chromosome \citep{lange_classical_2007}, RNA-interacting RG/RGG boxes for P-bodies and stress granules \citep{decker_p-bodies_2012}, and the aforementioned transportin RS repeats and GLFG nucleoporin repeats (regular expressions in Appendix~\ref{app:nls}).
Appendix~\ref{app:motifs} shows that RL-SAE installs the expected motif in nearly all generated sequences for five of the six motifs, substantially outperforming other methods. The exception is the P-body RG/RGG motif, which is only weakly present within the top enriched features (Appendix~\ref{app:galleries}).

\paragraph{Private feature selection reduces predicted off-target association}

\begin{figure}[h]\centering
\includegraphics[width=\textwidth]{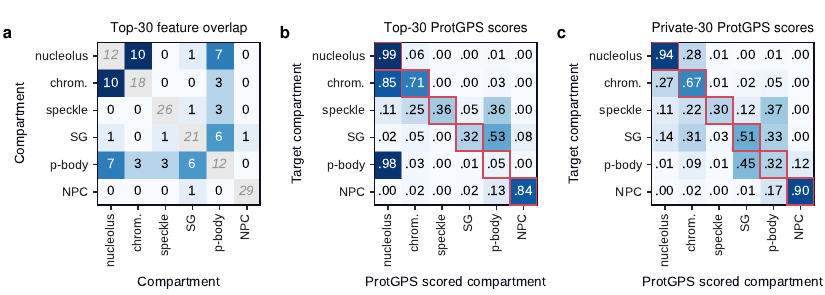}
\caption{
\textbf{Targeting private SAE features reduces predicted cross-talk between compartments.}
\textbf{(a)} Number of private (greyed diagonal) and shared (off-diagonal) features among the top-30 between the 6 considered compartments. \textbf{(b)} Heatmap of ProtGPS scores for sequences generated after RL-SAE on the top-30 features of each compartment. \textbf{(c)} Heatmap of ProtGPS scores of sequences generated after RL-SAE on the private-30 features of each compartment. 
}
\label{fig:selectivity}
\end{figure}

Our feature enrichment analysis lets us explicitly identify features shared between different sequence sets. Figure~\ref{fig:selectivity}a shows the number of shared top-30 features among the 6 biomolecular compartments. The nucleolus and chromosome share many enriched features, consistent with their nuclear localization and basicity. P-bodies share charged features with the nucleolus and aromatic features with stress granules, consistent with a hydrophobic sticker grammar in granule IDRs, and the NPC shares the fewest features, consistent with its unique FG-based grammar.
Figure~\ref{fig:selectivity}b shows that this feature cross-talk is reflected in the off-target ProtGPS scores. In particular, chromosome and P-body targeting sequences both score highly for the nucleolus, and stress granule targeting sequences score highly for P-bodies.

To test whether these shared features contribute to ProtGPS cross-talk, we run RL-SAE targeting the private-30 feature sets. We find that this substantially reduces off-target nucleolus scores for chromosome- and P-body-targeting sequences, and off-target P-body scores for stress-granule-targeting sequences (Figure~\ref{fig:selectivity}c). This reduced cross-talk occurs while maintaining strong on-target scores: private-30 sequences achieve higher mean on-target scores than SFT, activation steering, and natural IDRs across all six compartments (Table~\ref{tab:localization}). These experiments support the hypothesis that shared SAE features contribute to predicted cross-talk, and that feature selection can improve predicted compartment specificity. However, we also note that off-target scores are not uniformly reduced, and some off-target pairs increase.

\subsubsection{Designing and composing transcriptional effector domains}
\label{sec:res_private}

\paragraph{Designing activation and repression domains}
Our experiments on transcriptional activation and repression domains show similarly strong performance with RL-SAE. Using the PADDLE predictor to estimate activation strength, we show in Table \ref{tab:effector} that RL-SAE generates sequences with substantially higher PADDLE scores than those from SFT and steering. SFT, steering, and RL-SAE generate sequences with mean PADDLE scores approximately 89\%, 128\%, and 209\% of those of natural ADs, respectively. Inspection of generated sequences (Appendix~\ref{app:sequences}) reveals that both steering and RL-SAE generate canonical ADs with hydrophobic stickers interspersed within acidic stretches \citep{delrosso_large-scale_2023, staller_directed_2022}, justifying the strong predicted activation strength.

For repression, which is frequently driven by SUMO (Small Ubiquitin MOdifier) modification and subsequent corepressor recruitment \citep{delrosso_large-scale_2023}, we score sequences for the presence of $\psi$KxE SUMOylation consensus motifs and their negative charge-dependent (NDSM) variants \citep{yang_extended_2006} (regexes in Appendix~\ref{app:nls}). We find that RL-SAE generates over 5 motifs per sequence, compared to 0.60 per sequence for natural sequences and 0.56 per sequence for steered generations. 

SUMO modification is highly dynamic \citep{vertegaal_signalling_2022}, and interactions between SUMOylated proteins and their binding partners are weak and heavily dependent on avidity across multiple modified sites \citep{psakhye_protein_2012}. Thus, a greater density of SUMOylation motifs is expected to substantially increase corepressor recruitment, both by stabilizing the SUMO modification itself, and by enabling multivalent interactions with corepressors \citep{banani_compositional_2016}, as is observed in some of the strongest experimentally measured human repressor sequences \citep{delrosso_large-scale_2023}.

\begin{table}[h!]\centering
\footnotesize
\setlength{\tabcolsep}{3pt}
\begin{tabular}{llccccc}
\toprule
& \textbf{Readout} & Base & Nat. & SFT & Steering ($\alpha$) & RL-SAE \\
\midrule
\textit{Activation} & PADDLE & $1.03_{.03}$ & $5.34_{.18}$ & $4.75_{.08}$ & $\underline{6.81}_{.01}$ (0.43) & $\mathbf{11.16}_{.11}$ \\
\midrule
\textit{Repression} & $\psi$KxE & $0.07_{.02}$ & $\underline{0.60}_{.05}$ & $0.37_{.04}$ & $0.56_{.02}$ (0.35) & $\mathbf{5.63}_{.60}$ \\
 & NDSM & $0.01_{.01}$ & $0.10_{.02}$ & $0.06_{.00}$ & $\underline{0.21}_{.02}$ (0.35) & $\mathbf{1.80}_{.44}$ \\
\bottomrule
\end{tabular}
\caption{
\textbf{RL-SAE generates sequence features associated with transcriptional regulation.}
Values are PADDLE scores and densities of SUMOylation motifs per sequence. Bold indicates the best score per row, underline the second best. Motif regexes are in Appendix~\ref{app:nls}. Subscripted values are standard deviations over three seeds, except for natural IDRs (Nat.), which report the standard error of the mean (Appendix~\ref{app:stats}).
}
\label{tab:effector}
\end{table}

\begin{figure}[h!]\centering
\includegraphics[width=\textwidth]{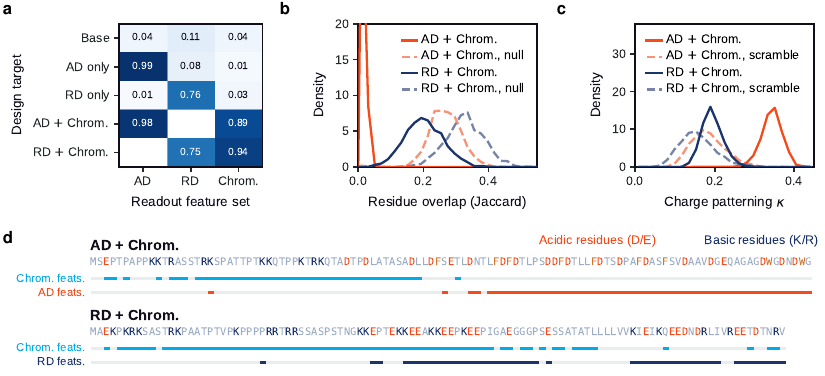}
\caption{
\textbf{Feature compatibility determines the organization of sequences with composed features.}
\textbf{(a)} Fraction of features activated in each target feature set by sequences generated under each RL-SAE condition. The chromosome feature set is private-30. 
\textbf{(b)} Distribution of the per-sequence Jaccard overlap between residues which fire on chromosome association features versus on activation domain or repression domain features. Curves show each design and a position-shuffled null. 
\textbf{(c)} Distribution of charge patterning $\kappa$ parameter for chromosome $+$ AD/RD, and scrambled variants. 
\textbf{(d)} Example generated chromosome $+$ AD/RD sequences showing feature firing positions for chromosome association (light blue) and the activation domain (AD, upper, red) or repression domain (RD, lower, dark blue).
}
\label{fig:conjunction}
\end{figure}

\paragraph{Composing chromosome association and effector features}
RL-SAE additionally allows us to compose arbitrary feature sets. To demonstrate this, we target the private-30 chromosome features together with the top-30 activation or repression domain features (chromosome $+$ AD/RD) to mimic a synthetic transcriptional regulator that must both localize to chromatin and possess gene regulatory function. The experiments demonstrate that RL-SAE effectively installs on average 53 of the 60 target features in generated sequences (Figure~\ref{fig:conjunction}a). Additionally, chromosome $+$ AD sequences retain PADDLE scores close to those of AD-only designs while also scoring highly for chromosome association and nuclear localization.
Chromosome $+$ RD sequences similarly achieve strong predicted chromosome association and nuclear localization, with multiple SUMOylation motifs per sequence (Table~\ref{tab:conjunction_main}).

\begin{table}[h!]\centering
\footnotesize
\setlength{\tabcolsep}{3pt}
\begin{tabular}{l c@{\hskip 8pt} cc@{\hskip 8pt} ccc}
\toprule
& \textbf{Activation} & \multicolumn{2}{c}{\textbf{Repression}} & \multicolumn{3}{c}{\textbf{Chromosome / nuclear}} \\
\cmidrule(lr){2-2}\cmidrule(lr){3-4}\cmidrule(lr){5-7}
\textbf{RL-SAE target} & PADDLE & $\psi$KxE & NDSM & ProtGPS & DeepLoc & cNLS \\
\midrule
chromosome\,$+$\,AD & \exq{$11.00_{.10}$} & $0.00_{.00}$ & $0.00_{.00}$ & \exq{$0.54_{.44}$} & \exq{$0.76_{.14}$} & \exq{$0.61_{.53}$} \\
chromosome\,$+$\,RD & $-0.09_{.05}$ & \exq{$2.89_{.10}$} & \exq{$1.73_{.77}$} & \exq{$0.73_{.14}$} & \exq{$0.92_{.02}$} & \exq{$0.99_{.02}$} \\
\bottomrule
\end{tabular}
\caption{\textbf{RL-SAE combines localization and transcriptional effector features.}
Evaluation of the two composed designs (chromosome $+$ AD/RD), each targeting nuclear localization together with one gene-regulatory function. Values are PADDLE scores, SUMOylation motif density, predicted localization, and nuclear localization signal presence (cNLS). Grey cells mark the values which are biologically expected to score highly. 
Subscripted values are standard deviations over three seeds (Appendix~\ref{app:stats}).}
\label{tab:conjunction_main}
\end{table}

\paragraph{Sequence organization reflects feature compatibility}
We find that how these features are installed depends on the compatibility of the two feature sets with one another.
For the chromosome $+$ AD set, we find that there is minimal Jaccard overlap between residues that fire for chromosome features and those that fire for activation domain features (Figure~\ref{fig:conjunction}b, red). Since chromosome associating features are basic and transcriptional activation features are acidic, the model learns to segregate them into two distinct charge blocks (example sequence in Figure~\ref{fig:conjunction}d, upper), resulting in substantially greater charge blockiness ($\kappa$) for chromosome $+$ AD sequences versus their scrambles (Figure~\ref{fig:conjunction}c, red).
On the other hand, substantially more residue overlap is observed between chromosome and repressor domain features: the features which are associated with transcriptional repression are partly basic and the model finds a way to partially overlap those motifs with basic chromosome features in the generated sequences, resulting in greater Jaccard overlap and lower $\kappa$ (Figure~\ref{fig:conjunction}b,c, blue). Together, these results suggest that RL-SAE can compose distinct grammars into generated sequences, with the model learning how to organize features to accommodate their composition and patterning requirements without explicit constraints.

\section{Conclusions}

We introduce \textbf{IDiom}, a protein language model for the generative design of IDRs, and \textbf{reinforcement learning with sparse autoencoder features (RL-SAE)}, a post-training method that enables a generative model to activate specified sets of SAE features within generated sequences. Our experiments demonstrate that IDiom generates diverse sequences which recapitulate the properties of natural IDRs, while RL-SAE effectively installs tens of target features and produces sequences with strong predicted biological function. Together, IDiom and RL-SAE provide a framework for identifying and composing interpretable sequence features for IDR design, and future experimental validation could establish their utility for engineering biological function. More broadly, RL-SAE could enable the composable design of proteins by jointly targeting interpretable features, such as those associated with catalysis, binding, or stability.

\paragraph{Limitations}

One limitation of RL-SAE is that it requires training, and scanning across many combinations of feature sets could be computationally intensive. 
Also, our top-$N$ feature selection is simplistic and may omit functionally relevant features or include redundant ones.
Finally, the evaluations in this paper are computational and experimental validation is needed to confirm whether the predicted properties translate into biological function.

\section*{Reproducibility statement}

Code can be found at: \url{https://github.com/rotskoff-group/idiom} \\
Models and data can be found at: \\
\url{https://huggingface.co/collections/jxliu2/idiom}.

Dataset curation is described in Appendices~\ref{app:corpus} and~\ref{app:labels}. Model and training details are described in Appendices~\ref{app:model}, \ref{app:sae}, and~\ref{app:rl}. Feature selection is described in Appendix~\ref{app:signatures}. Baseline methods are described in Appendices~\ref{app:rl}, \ref{app:baselines}, and \ref{app:rlgps}. Evaluation procedures are provided in Appendices~\ref{app:generation}, ~\ref{app:disorder}--\ref{app:nls}, and \ref{app:stats}.

\section*{Ethics statement}

This work develops a computational method which enables the generative modeling of intrinsically disordered protein regions. The evaluations in this work are computational, and generated sequences must undergo appropriate experimental validation and biosafety assessment before future use. In addition, such protein sequence generation methods may be repurposed for harmful use, and this potential should be considered when distributing and applying these methods. We confirm that we comply with the licenses and attribution requirements of all datasets used in this work.

\section*{AI use statement}

In this work, we used generative AI tools to assist in finding relevant literature sources, to aid in editing manuscript text, and to help implement research code. We extensively revised all AI-edited text and manually verified all AI-implemented code. We take full responsibility for the final content of this work, including text, claims or artifacts produced with the aid of generative AI.

\section*{Acknowledgments}

J.X.L. acknowledges support from the NIH T32, award number T32HL094274. S.I. acknowledges support from the NSF GRFP and the Shoucheng Zhang Graduate Fellowship. F.H. acknowledges support from the Stanford Center for Molecular Analysis and Design Fellowship. This research was supported by the National Institute of General Medical Sciences of the NIH under award number 1R35GM159834-01 (G.M.R.) and 5R35GM130332 (A.R.D.). The content is solely the responsibility of the authors and does not necessarily represent the official views of the NIH. The authors acknowledge the use of the Stanford Sherlock compute cluster.

\bibliography{references}
\bibliographystyle{iclr2026_conference}

\newpage
\appendix
\raggedbottom  %

\section*{Appendix}

\section{IDiom-DB construction}\label{app:corpus}

\begin{figure}[H]\centering
\includegraphics[width=\textwidth]{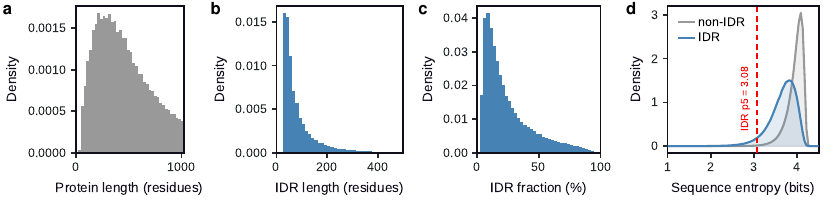}
\caption{
\textbf{Characterization of IDiom-DB.} 
\textbf{(a)} Length distribution of full-length protein sequences within which the IDRs are located, of the final curated dataset. 
\textbf{(b)} Length distribution of IDRs within the final curated dataset. 
\textbf{(c)} Distribution of the fraction of residues of the full-length sequence which are IDR residues. 
\textbf{(d)} Distribution of sequence entropy for IDR versus non-IDR regions in the final curated dataset. The vertical red dashed line indicates the sequence entropy threshold at the lower 5th percentile. 
}\label{fig:si_corpus}
\end{figure}

\paragraph{Data curation}
We curate IDiom-DB from the AlphaFold Database (AFDB), version 4. First, we use MMseqs2 to cluster the 214M AFDB sequences at 90\% identity and 80\% coverage (other MMseqs2 parameters below). Next, we follow the method of \citet{tesei_conformational_2024} to determine the locations of pLDDT-based IDRs within the MMseqs2 cluster representative proteins. In this method, we first apply a 15 residue-wide averaging filter to the pLDDT values. Next, we mark residues with $\mathrm{pLDDT} > 80$ as folded, $\mathrm{pLDDT} < 70$ as disordered, and $70 \leq \mathrm{pLDDT} \leq 80$ as gap regions. Folded and disordered regions with a length shorter than ten residues are reclassified as gaps. If a gap region is flanked by two disordered regions, or if it is N- or C-terminal and is adjacent to a disordered region, we relabel it as disordered. All other gap regions are relabeled as folded. Each record within the dataset corresponds to a different IDR, and any given protein may yield $\ge1$ IDR. IDRs which are located in proteins whose full length is greater than 1,020 residues, IDRs shorter than 30 residues long, and sequences whose entire length is low-pLDDT are discarded. Finally, we deduplicate DisProt sequences from this dataset and filter out signal peptides (both described below). This curation process yields 54M IDRs and their associated N- and C-terminal flanking contexts.

\paragraph{Sequence clustering and percent identity characterization}
We use MMseqs2 to cluster sequences in the AlphaFold Database at 90\% identity and 80\% coverage. MMseqs2 is run using the following command: \texttt{mmseqs linclust --min-seq-id 0.9 --cov-mode 0 -c 0.8 --cluster-mode 2}.

We also use MMseqs2 to determine the sequence identity of generated sequences with respect to the training set. Specifically, we compare prompted and unprompted IDR sequences against the 54M IDRs in IDiom-DB (IDR vs IDR only). MMseqs2 is run with the following command to search the generated sequences against the IDiom-DB IDRs: \texttt{mmseqs search --max-seqs 1 -e 1e3 --min-seq-id 0.0 -c 0.0}. For each generated sequence, this command returns the sequence identity of the closest match in the training set, whose distribution is plotted in Figure~\ref{fig:model}f.

\paragraph{DisProt deduplication}
We use the experimentally validated DisProt IDRs as a ground truth IDR reference for both dataset curation as well as for evaluating our generated sequences. To prevent leakage of these sequences into the training data, we remove IDRs from the curated training dataset which have high sequence similarity to a DisProt IDR. DisProt IDRs are first extracted using experimentally annotated disordered regions (consensus structural state "D") \citep{nugnes_disprot_2026}. IDRs which are at least 30 residues long, which reside in proteins whose full length is $\le$ 1,020 residues, and which are not 100\% disordered are retained, yielding 1,663 benchmark IDRs. We next use MMseqs2 to search these DisProt IDRs against the IDR spans of the 54M curated training IDRs, and we discard any training record whose IDR aligns with a DisProt IDR with at least 50\% sequence identity and 80\% bidirectional coverage (\texttt{--cov-mode 0}). MMseqs2 is run using the command: \texttt{mmseqs search --min-seq-id 0.5 -c 0.8 --cov-mode 0 -s 7 --max-seqs 300}, and this procedure removes 55K records.

\paragraph{Signal peptide filtering}
Signal peptides are short N-terminal sequences which are often cleaved after protein expression and subsequently degraded. Residues corresponding to N-terminal signal peptides often have low AF2 pLDDT scores and are disordered in solution. However, they are cleaved and thus absent from the fully mature proteins, so we do not consider them as suitable IDRs for our dataset \citep{piao_co-attention_2026}. We therefore use the SignalP 6.0 predictor to identify potential signal peptide cleavage sites for filtering \citep{teufel_signalp_2022}. We consider N-terminal subsequences to be signal peptides if they precede a SignalP cleavage residue that has a predicted probability $\ge$ 0.9. If an identified IDR contains such a signal peptide, we trim those residues and keep the remainder of the IDR. If the trimmed IDR length falls below 30 residues, then we discard the entry entirely. This process removes 3.6M records, yielding the final set of 54M IDRs.

\paragraph{Tokenization}
We tokenize protein sequences using a simple alphabet in which each amino acid is represented by a single token. We introduce the tokens \texttt{<N>}, \texttt{<C>}, and \texttt{<I>} to denote the N-terminal flanking context of an IDR, the C-terminal flanking context, and the IDR span itself, respectively. We additionally add the standard beginning-of-sequence \texttt{<bos>} and end-of-sequence \texttt{<eos>} tokens to all sequences, and we pad all sequences to the maximum length of 1,024 tokens using the \texttt{<pad>} token. The total size of the alphabet we use is 27 tokens (20 amino acids, \texttt{<N>}, \texttt{<C>}, and \texttt{<I>}, and \texttt{<bos>}, \texttt{<eos>}, \texttt{<pad>}, and \texttt{<mask>}).

\paragraph{Data splitting and augmentation}
To process IDiom-DB for model pretraining, we first randomly split the 54M IDR records into 99\% training, 0.5\% validation, and 0.5\% test sets. Next, we transform the protein sequences into a fill-in-the-middle format. We prepend the token \texttt{<N>} to any N-terminal context of the IDR, prepend \texttt{<I>} to the IDR itself, and prepend \texttt{<C>} to any C-terminal context of the IDR. We then rearrange the sequence in the order \texttt{<N><N-terminal context><C><C-terminal context><I><IDR span>}. An example sequence is \texttt{<N>MEDS...HLVA<C>SVED...RKSL<I>VEED...KGPS}.

During training, each sampled record is presented with its N- and C-terminal flanking contexts (prompted) with 50\% probability and without flanks (unprompted) with 50\% probability. An example unprompted sequence is \texttt{<N><C><I>VEED...KGPS}.

\newpage

\section{Model architecture, pretraining, and inference}\label{app:model}

\paragraph{Model architecture}
IDiom is a decoder-only transformer which we train in three model sizes: 19M (6 layers, hidden dimension 512), 85M (12 layers, hidden dimension 768), and 302M parameters (24 layers, hidden dimension 1,024). The model employs pre-normalization with RMSNorm, applies query-key normalization within each attention block, and uses the SwiGLU non-linearity \citep{shazeer_glu_2020} in all feedforward networks. The feedforward expansion ratio is 8/3, and positional information is encoded using Rotary Position Embeddings (RoPE) \citep{su_roformer_2023}. The model processes sequences with a maximum length of 1,024 tokens, and shorter sequences are padded to this length. Multi-head attention is calculated using Flash Attention \citep{dao_flashattention_2022}.

\paragraph{Pretraining}
We pretrain IDiom on the training split of IDiom-DB using the prompted/unprompted sampling procedure described in Appendix~\ref{app:corpus}. All sequences are padded to a fixed length of 1,024 tokens. The model is trained autoregressively using next-token prediction.

Training is performed using the AdamW optimizer and no weight decay. Learning rates of $4.0 \times 10^{-4}$, $3.5 \times 10^{-4}$, and $3.0 \times 10^{-4}$ are used for the 19M, 85M, and 302M parameter models, respectively.
The learning rate schedule uses a linear warmup over the first 3,000 steps, followed by cosine annealing decay to a minimum of 10\% the initial learning rate over 250,000 total training steps. The global batch size is 1,024 (Distributed Data Parallel training). We use a cross-entropy loss while ignoring contributions from pad or mask tokens. Training is performed in mixed precision (fp32/bfloat16) and with gradient clipping at 1.0. Model validation is performed every 25,000 training steps on the validation split, and training is run for 250,000 optimizer steps.

\paragraph{Model inference}
We generate IDR sequences from IDiom using autoregressive decoding. At each position $t$ in the sequence, we calculate the model logits $z_t$ and convert them to a probability distribution via $p(x_t | x_{<t}) = \text{softmax}(z_t / T)$, then sample the next token from this full categorical distribution over the vocabulary. We use a fixed sampling temperature of $T = 1.0$ for all generations. 

Sequence generation supports both prompted and unprompted modes. For prompted generation, the N- and C-terminal flanking contexts are provided as the prompt in fill-in-the-middle format: \texttt{<N><N-terminal context><C><C-terminal context><I>}, and the model generates the IDR span autoregressively following the \texttt{<I>} token. For unprompted generation, the prompt consists of only the three special tokens without any flanking context: \texttt{<N><C><I>}, and the model generates the disordered sequence. In both cases, generation terminates upon sampling the \texttt{<eos>} token or upon reaching the maximum sequence length of 1,024 tokens.

\newpage

\section{Characterization of generated sequences}\label{app:generation}

\begin{figure}[H]\centering
\includegraphics[width=\textwidth]{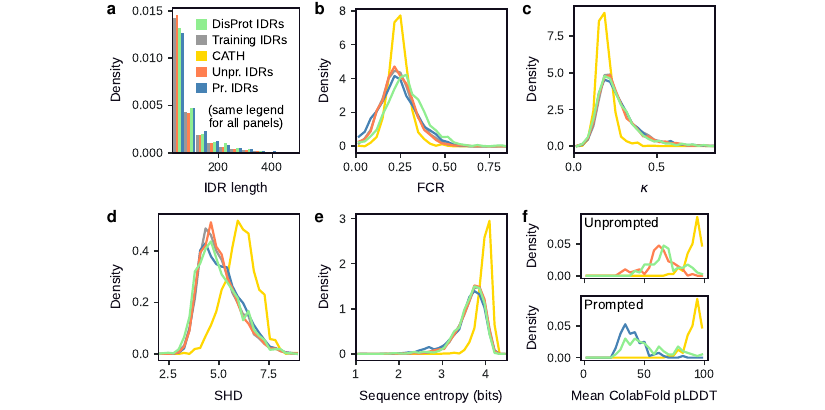}
\caption{
\textbf{Characterization of training IDRs, IDiom-generated IDRs, DisProt IDRs, and CATH sequences.} 
\textbf{(a)} Distribution of training, generated, and DisProt IDR lengths. The legend in this panel applies to all other panels in this figure. 
\textbf{(b)} Distribution of the fraction of charged residues (FCR) for training, generated, and DisProt IDRs, and CATH sequences. 
\textbf{(c)} Distribution of charge patterning $\kappa$ parameter. 
\textbf{(d)} Distribution of sequence hydropathy decoration (SHD). 
\textbf{(e)} Distribution of sequence entropy. 
\textbf{(f)} Distribution of AlphaFold2 scores for CATH sequences, IDiom-generated IDRs, unprompted (upper) and prompted (lower), as well as DisProt IDRs without surrounding context (upper) and with context (lower). 
}
\label{fig:si_generation}
\end{figure}

\paragraph{Generated sequences}
To characterize the pretrained 302M parameter IDiom model, we generate 100,000 unprompted IDRs and 166,300 IDRs conditioned on the N- and C-terminal flanking contexts of 1,663 DisProt IDRs (100 IDRs per protein).

\paragraph{Amino acid composition}
To analyze the compositional biases of the generated IDRs, we calculate the amino acid enrichment of training, generated, and DisProt sequences relative to a baseline composition of the folded CATH domains (Figure~\ref{fig:model}e). Consistent with established compositional biases of disordered regions \citep{theillet_alphabet_2013, ruff_molecular_2026}, generated sequences are strongly enriched in proline and serine, and depleted in order-promoting aliphatics such as leucine, isoleucine, and valine, as well as the aromatics phenylalanine, tryptophan, and tyrosine. The trends for training and generated sequences closely match natural DisProt IDRs across most amino acids, with the closest agreement observed for IDRs generated with the DisProt flanking contexts as prompts. We note that the discrepancies with DisProt in the generated compositions of cysteine and charged residues (glutamic acid, lysine, and aspartic acid) may be attributed to the small size and selection bias of the DisProt dataset.

\paragraph{Sequence patterning}
IDRs exhibit sequence patterning features that differ substantially from those of folded domains, including characteristic charge distributions, hydrophobic residue patterning, and low-complexity compositions \citep{langstein-skora_sequence_2026, ruff_molecular_2026, das_relating_2015}. IDiom-generated sequences illustrate canonical IDR features such as Q/N-rich low-complexity regions \citep{van_der_lee_classification_2014}, polyampholyte charge block patterning \citep{mitrea_self-interaction_2018}, prion-like aromatic/glycine patterning \citep{martin_valence_2020}, and proline enrichment \citep{theillet_alphabet_2013}. To assess how well IDiom recapitulates these properties, we calculate the distributions of four sequence-level descriptors (discussed below) for the generated, training, and DisProt disordered sequences, as well as for the folded CATH domain sequences (Figure \ref{fig:si_generation}b-e), together with the generated IDR length distribution (Figure \ref{fig:si_generation}a) and AF2 pLDDTs (Figure \ref{fig:si_generation}f). 

\paragraph{Charge content}
Electrostatic interactions strongly influence the conformational behavior of IDRs, and both the overall charge content and its linear patterning are closely linked to physical properties and biological function \citep{mantonico_acidic_2024, lin_charge_2017, mitrea_self-interaction_2018, das_relating_2015}. The fraction of charged residues (FCR, Figure \ref{fig:si_generation}b) distinguishes strongly charged polyampholytes from weakly charged sequences. We find that the generated and natural IDRs span a wider range of FCR values than folded CATH domains, reflecting the heterogeneity of natural IDRs, which range from highly charged sequences in which electrostatic repulsion drives disorder, to weakly charged low-complexity sequences \citep{das_conformations_2013}.

\paragraph{Charge patterning}
To characterize charge patterning, we calculate the linear charge patterning parameter $\kappa$, which quantifies the deviation of a given sequence from a maximally charge segregated permutation of the same sequence \citep{das_conformations_2013}. Values of $\kappa \approx 0$ indicate well-mixed opposite charges and $\kappa \approx 1$ indicates segregation into blocks of the same charge, and we plot these distributions in Figure \ref{fig:si_generation}c. Consistent with prior work linking charge segregation in IDRs to intermolecular interactions and phase behavior \citep{mitrea_self-interaction_2018, ginell_sequence-based_2025}, natural IDRs from the training and DisProt sets exhibit a tail toward high $\kappa$ values relative to CATH domains, a feature that IDiom reproduces in its generated sequences.

\paragraph{Hydrophobic patterning}
We next examine hydrophobic patterning using the sequence hydropathy decoration (SHD) metric, which quantifies the spatial clustering of hydrophobic residues along the chain \citep{zheng_hydropathy_2020}. Generated sequences exhibit substantially lower SHD values than folded CATH domains (Figure \ref{fig:si_generation}d), consistent with the reduced hydrophobic clustering in disordered regions that prevents hydrophobic collapse \citep{zheng_hydropathy_2020}. This trend closely matches natural DisProt and training set IDRs, and it contrasts with folded CATH domains, whose higher SHD values reflect the locally concentrated hydrophobic residues required to stabilize buried protein cores \citep{dyson_role_2006}.

\paragraph{Sequence entropy}
Finally, we calculate each sequence's entropy as the Shannon entropy (bits) of its amino acid composition. IDRs frequently contain low-complexity segments \citep{romero_sequence_2001}, and natural DisProt and training set IDRs show lower entropy than folded CATH domains. Prompted and unprompted IDRs closely reproduce this shift, with the entropy of generated sequences matching the DisProt distribution well (Figure \ref{fig:si_generation}e). Taken together, these results show that IDiom has learned the sequence grammar of disordered regions across multiple descriptors, and that generated sequences recapitulate the sequence patterning features of natural IDRs.

\paragraph{Structure prediction}
We use ColabFold \citep{mirdita_colabfold_2022} to perform AlphaFold2 structure predictions on generated sequences and determine the predicted local distance difference test (pLDDT) values.

\paragraph{Sequence analysis}
We use the Sparrow package \citep{holehouse_sparrow_2022} to calculate the fraction of charged residues, the linear charge patterning parameter $\kappa$, and sequence hydropathy decoration.

\newpage

\section{Sparse autoencoder training and characterization}\label{app:sae}

\begin{figure}[H]\centering
\includegraphics[width=0.25\textwidth]{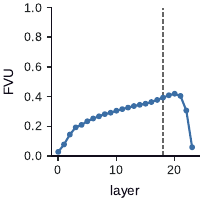}
\caption{
\textbf{Sparse autoencoder layer sweep.}
Fraction of variance unexplained (FVU) per layer, evaluated on 10K held-out validation records
($k=32$, expansion $\times16$).
The layer-18 is marked by the vertical dashed line.
}
\label{fig:si_sae}
\end{figure}

\paragraph{Sparse autoencoder training}
We train TopK sparse autoencoders \citep{cunningham_sparse_2023, gao_scaling_2024}, IDiomSAEs, on the residual stream of all layers of the base IDiom model:
\begin{equation*}
\vz = \operatorname{TopK}\!\big(\operatorname{ReLU}(\mW_{\text{enc}}(\vx - \vb_{\text{dec}}) + \vb_{\text{enc}})\big), \qquad \hat{\vx} = \mW_{\text{dec}}\, \vz + \vb_{\text{dec}}
\end{equation*}
with $k = 32$ active latents from $16{,}384$ total ($16\times$ expansion factor), and where $\vb_{\text{dec}}$ is initialized from the mean of the training activations. The model is trained to minimize the fraction of variance unexplained: $\mathcal{L}_{\text{FVU}} = \|\vx - \hat{\vx}\|^2 / \|\vx - \bar{\vx}\|^2$, with an auxiliary AuxK loss (with weight $0.03125$) to revive dead latents \citep{gao_scaling_2024}. Training activations are taken only from IDR residue positions, and only unprompted IDRs are used for training. IDiomSAEs are trained using Adam with a learning rate of $2 \times 10^{-4}$ and batch size of 4,096, for 50,000 optimizer steps. We use a 1,000 step warmup. For all analysis in this paper, we use the layer-18 IDiomSAE.

\newpage

\section{Feature enrichment procedure}
\label{app:signatures}

\paragraph{Significance testing}
To determine the SAE features which are most enriched in a set of positively labeled sequences $\mathcal{P}$ relative to a set of negatively labeled sequences $\mathcal{N}$, we follow the method of \citet{candido_language_2026}. For each feature $j$ of the 16,384 SAE features, we set $a_j$ and $b_j$ to be the number of sequences in $\mathcal{P}$ and $\mathcal{N}$, respectively, in which $j$ fires, $K_j = a_j + b_j$ to be the total firing count pooled across both sets, and $M = |\mathcal{P}| + |\mathcal{N}|$ to be the total pooled sequence count. Under the null hypothesis that firing is independent of $\mathcal{P}$ or $\mathcal{N}$ membership, drawing $|\mathcal{P}|$ sequences without replacement from the pooled $M$ (of which $K_j$ fire) gives a firing count that is hypergeometrically distributed with mean $\mu_j$ and variance $\sigma_j^2$. We standardize the observed count $a_j$ against this null:
\begin{equation}
\mu_j = \frac{|\mathcal{P}|\,K_j}{M}, \qquad
\sigma^2_j = \frac{K_j\,(M - K_j)\,|\mathcal{P}|\,(M - |\mathcal{P}|)}{M^2\,(M-1)}, \qquad
Z_j = \frac{a_j - \mu_j}{\sqrt{\sigma^2_j}}.
\end{equation}
We then calculate a two-sided $p$-value from $Z_j$ and control the false discovery rate with the Benjamini-Hochberg procedure. The log odds ratio (LOR, Equation~\ref{eq:lor}) is calculated with a $0.5$ pseudocount added to all four contingency table cells.

\paragraph{Feature filtering and ranking}
We discard features whose peak activating position lies within 2 residues of either edge of an IDR in $\ge$ 50\% of feature firings, since these are likely artifacts from the process of extracting IDRs from their full sequence context. Then, we filter out features with a false discovery rate of $>10^{-3}$, similarly remove features with LOR $\le 1$, and only consider features $j$ which fire in at least 5\% of sequences in the positive set $\mathcal{P}$.
We rank the remaining features by LOR and take the top-$N$ as the feature set for analysis. All results in the main text use $N=30$.

\newpage

\section{Compartment and effector domain sequences}\label{app:labels}

\paragraph{Biomolecular compartments}
To curate a positive set $\mathcal{P}$ of IDR sequences with annotated localization to various biomolecular compartments, we start with the dataset curated by \citet{kilgore_protein_2025}. That dataset consists of a curated set of human full-length protein sequences with UniProt or CD-CODE annotated association with 12 biomolecular condensates. We further process the dataset by first selecting six compartments for analysis: nucleolus, chromosome, stress granule, P-body, nuclear speckle, and the nuclear pore complex (NPC), and then removing sequences with annotated association with more than one compartment. Next, we run the Metapredict V3 disorder predictor on these sequences to isolate the IDRs and only keep IDRs with length 30--1,020 residues. We then use these sequences as the $\mathcal{P}$ set. The $\mathcal{N}$ set is 100,000 length-matched IDRs randomly selected from IDiom-DB.

\paragraph{Effector domains}
To curate a positive set $\mathcal{P}$ of human transcriptional effector domain sequences, we start with a dataset of 80-residue sequences from a high-throughput tiling screen \citep{delrosso_large-scale_2023}. For both the activation domain and repression domain datasets, we apply the same processing. We first merge overlapping hit sequences to avoid duplication. Next, we run the Metapredict V3 disorder predictor on the full-length parent proteins of the sequences and keep only sequences where at least 50\% of residues are predicted disordered. This yields 288 AD sequences and 408 RD sequences. For the negative set, we use sequences which are non-hits in all three experimental screens (ON+OFF counts $>$ 50 and average score below the control mean $+$ 2 SD), without merging overlaps, and with the same Metapredict V3 IDR filtering. This yields 4,920 inert tiles as the $\mathcal{N}$ set.

\newpage

\section{RL-SAE details}\label{app:rl}

\begin{figure}[H]\centering
\includegraphics[width=\textwidth]{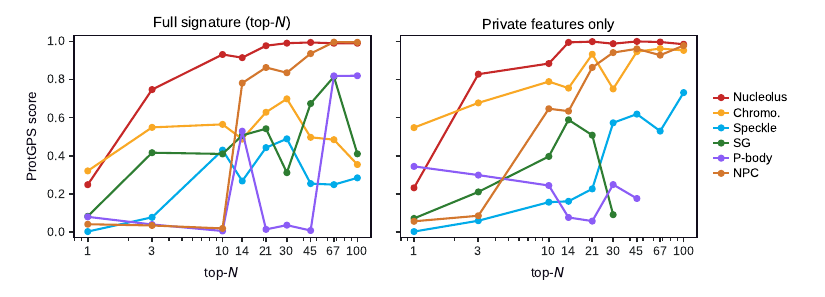}
\caption{
\textbf{ProtGPS dose-response against top-$N$.}
Left: Average ProtGPS score for sequences generated after RL-SAE on the top-$N$ features per compartment. Right: Average ProtGPS score for sequences generated after RL-SAE on the top-$N$ features which are private to the given compartment. Stress granule and P-body only have 36 and 55 private features, respectively.
}
\label{fig:si_dose}
\end{figure}

\begin{figure}[H]\centering
\includegraphics[width=\textwidth]{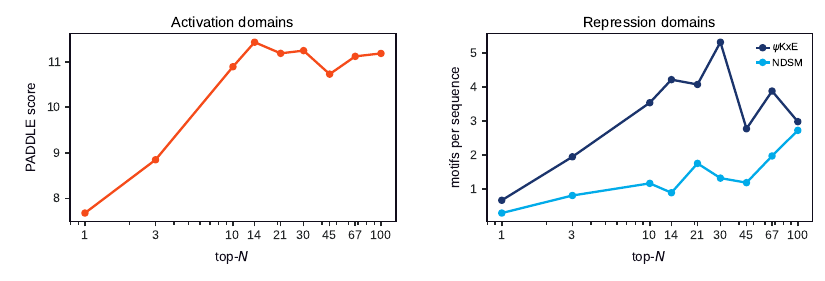}
\caption{
\textbf{Transcriptional effector domain dose-response against top-$N$.}
Left: PADDLE score for activation domains generated after RL-SAE on the top-$N$ activation domain features. Right: $\psi$KxE and NDSM motif density in sequences generated after RL-SAE on the top-$N$ repression domain features. 
}
\label{fig:si_effector_dose}
\end{figure}

\paragraph{Post-training}
We post-train IDiom using group relative policy optimization (GRPO) \citep{shao_deepseekmath_2024}. During GRPO training, for each batch of sequence prompts, the model generates multiple IDRs per prompt. Rewards are calculated for each generated sequence using the reward function defined in Equation~\ref{eq:reward}.
For RL-SAE, feature activations are computed using the frozen pretrained IDiom model and IDiomSAE. 
When performing RL on the ProtGPS oracle baseline, $R_S(y)$ is replaced with the ProtGPS score (value between 0 and 1) for the target compartment.

Advantages are calculated as normalized relative rewards per group: $A_i = \frac{r_i - \bar{r}_g}{\sigma_g}$, where $\bar{r}_g$ and $\sigma_g$ are the group-wise mean and standard deviation. The GRPO loss combines PPO-style clipping ($\epsilon_{\text{clip}}=0.2$) with a KL penalty ($\beta_{\text{KL}}=0.02$), applied per-token on completions only. We use a learning rate of $5\times10^{-6}$ with AdamW, group size of 8, and global batch size of 8. Training is performed for 3,000 steps, and all runs generate unprompted IDRs with a temperature of 1.0.

\paragraph{Top-$N$ score dependence}
In Figures~\ref{fig:si_dose} and \ref{fig:si_effector_dose} we plot the ProtGPS, PADDLE, and repression domain motif scores as a function of top-$N$ features on which RL-SAE is applied. Across most metrics, we observe that there is generally an upward trend of score versus top-$N$, but with substantial variability in ProtGPS scores. For all experiments in the main text, we choose the top-30 features for analysis because it is the largest $N$ on our sweep grid at which all six compartments have sufficient private features (Figure~\ref{fig:si_dose}, right panel).

\newpage

\section{Baseline methods}\label{app:baselines}

\paragraph{Activation steering}
For the same target feature sets as in RL-SAE, we sum and normalize the decoder vectors $\mW_{\mathrm{dec}}$ into the steering direction $\hat{\vd}_S$:
\begin{equation}
\vd_S = \sum_{j \in S} \mW_{\mathrm{dec}}[:,j], \qquad
\hat{\vd}_S = \frac{\vd_S}{\|\vd_S\|_2},
\end{equation}
where $\mW_{\mathrm{dec}}[:,j]$ is the decoder column corresponding to feature $j$. At every token position $t$ during inference, we modify the residual stream vector $\vx_t$ at layer-18 of IDiom by adding $\hat{\vd}_S$ scaled by some fraction $\alpha$ of the original residual stream vector's magnitude before renormalizing:
\begin{equation}
\tilde{\vx}_t = \vx_t + \alpha \|\vx_t\|_2\, \hat{\vd}_S, \qquad
\vx_t' = \|\vx_t\|_2 \, \frac{\tilde{\vx}_t}{\|\tilde{\vx}_t\|_2}
\end{equation}
and then substitute $\vx_t$ with $\vx_t'$ in the forward pass.

We steer using the same top-$N$ feature sets as RL-SAE, sweeping 40 values of $\alpha$ between 0 (base model) and 1.5 which are spaced more densely at low $\alpha$. Sequences are generated unprompted at a temperature of 1.0. We only keep generated sequences 50--150 residues in length, and we oversample to generate the required number of sequences. We discard any $\alpha$ whose generated sequences have a mean sequence entropy below the lower 5th percentile of IDR sequence entropies in IDiom-DB (3.08 bits), and we report steering results at the remaining $\alpha$ (parenthesized in tables) which achieves the highest score.

\paragraph{Supervised fine-tuning}
For each target, we fine-tune the base checkpoint on that target's positive set $\mathcal{P}$ (Appendix~\ref{app:labels}), holding out 10\% of sequences for validation, and using the unprompted mode with a completion-only loss. We train with AdamW with a learning rate of $1 \times 10^{-5}$, weight decay $0.1$, gradient clipping at 1.0, with a batch size of 16, for 1000 optimizer steps (number of epochs ranges from roughly 21 to 178 depending on the size of $\mathcal{P}$), with a 100 step warmup and then cosine decay.

\paragraph{Oracle RL baseline}
For biomolecular compartment association, we use the ProtGPS score (0 to 1) as the baseline reward in lieu of $R_S(y)$ in Equation~\ref{eq:reward}. ProtGPS is trained using the full-length protein sequences of the same per-compartment positive sequence sets. We refer to these runs as RL-GPS, and all hyperparameters are otherwise the same as in RL-SAE.

\newpage

\section{Reported uncertainties}\label{app:stats}

\paragraph{Variability across seeds}
For the three-seed table comparisons, we report the mean and standard deviation of a score across three seeds for each generative method. For RL-SAE, RL-GPS, and SFT, the three seeds are independent training runs with different random seeds. We score 1000 sequences generated from each final checkpoint, calculate a per-run average, and report the mean and standard deviation of the three per-run averages. For the base model and activation steering (no training), the three seeds are independent generation seeds (of 1000 sequences), and we report the mean and standard deviation across the three generated sets in the same way. The scrambled control is only done for RL-SAE and is calculated in the same way. Scores are calculated over 1000 generated sequences for all metrics except DeepLoc predictions, for which we use a 200 sequence subset due to computational cost. 
For natural IDR sets we report the mean and its standard error over all sequences in the set.

\newpage

\section{Top-$N$ coverage}\label{app:coverage}

\begin{figure}[H]\centering
\includegraphics[width=\textwidth]{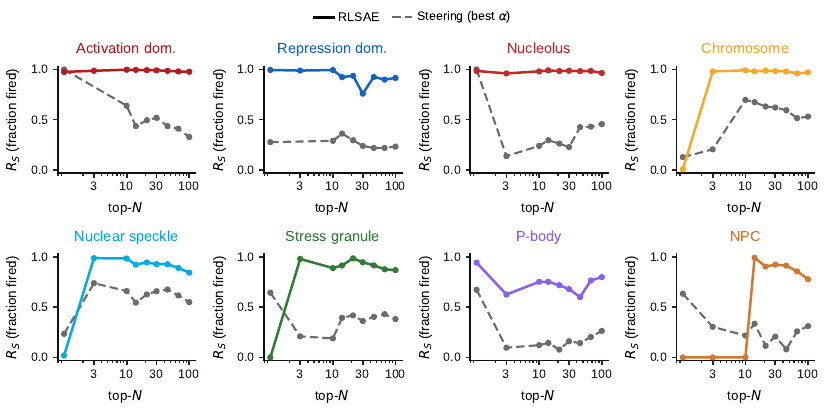}
\caption{
\textbf{Feature activation fraction versus top-$N$.} 
For all 8 sequence sets (2 transcriptional effector sequences and 6 compartment association sequences) the average fraction of target features which fire within generated sequences, $R_S$, as a function of top-$N$, for RL-SAE (solid colored lines) and activation steering (grey dashed lines). $R_S$ for steering is presented at the best $\alpha$ value across the sweep, regardless of whether that $\alpha$ passes the sequence entropy threshold (i.e. steering $R_S$ values presented here are a generous upper bound).
}
\label{fig:si_coverage}
\end{figure}

\paragraph{Target feature coverage}
In Figure~\ref{fig:si_coverage}, we plot the average fraction of the target features which fire within generated sequences, $R_S(y)$ (Equation~\ref{eq:reward}), as a function of top-$N$ for RL-SAE versus steering. For this comparison, we plot the best $R_S(y)$ achievable by steering, regardless of whether the corresponding $\alpha$ passes the sequence entropy threshold.

We find that when steering, for 5 of 8 sequence sets (activation domains, nucleolus, stress granule, P-body, NPC), $R_S(y)$ starts at a high value at $N=1$ and drops as $N$ is increased. When steering towards a number of directions beyond $N=1$, the activations are pushed towards the centroid of the set of directions. We hypothesize that as $N$ grows, this steered centroid direction becomes less representative of any individual feature's direction, hence the lower values of $R_S(y)$ as $N$ is increased. 

On the other hand, we observe that RL-SAE is able to effectively attain $R_S(y)$ values near 1.0 across most compartments and most $N$ values. The exceptions we find are several low-$N$ cases, such as $N=1$ for chromosome, nuclear speckle, and stress granule, and $N=1,3,10$ for NPC. Potentially, this is due to the rarity of some of these few top features: these features may be infrequently sampled by the base model and are thus difficult to optimize for.

\newpage 

\begin{figure}[H]\centering
\includegraphics[width=\textwidth]{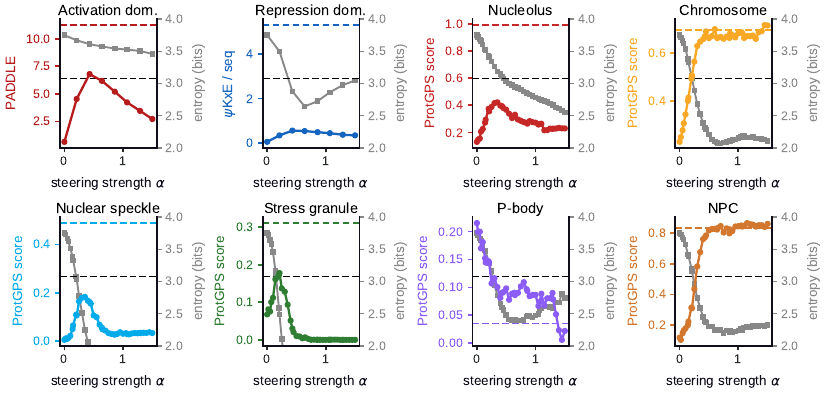}
\caption{
\textbf{Steering dose-response at $N\!=\!30$.} 
Here, we plot the various scoring metrics (PADDLE, $\psi$KxE, NDSM, and ProtGPS) versus activation steering strength $\alpha$ at top-$N=30$. The metrics are plotted as colored curves (left axis), and the generated sequence entropies are plotted as grey curves (right axis). 
}
\label{fig:si_steer_dose}
\end{figure}

\paragraph{Steering strength curves}
In Figure~\ref{fig:si_steer_dose}, we plot various scoring metrics (PADDLE, $\psi$KxE density, NDSM density, ProtGPS score) for sequences generated from the activation steering baseline, as a function of the steering strength $\alpha$.  The most common trend we observe (5 of 8 cases: activation domains, repression domains, nucleolus, nuclear speckle, and stress granule) is an initial increase in score as a function of $\alpha$, up to a peak value at an $\alpha$ between 0.2 and 0.45, followed by a decrease in the score at larger values of $\alpha$. This is accompanied by a concomitant reduction in sequence entropy with increased $\alpha$. Interestingly, we find that for 4 sequence sets (repression, nucleolus, nuclear speckle, and stress granule), the peak and subsequent drop in score occurs at approximately the same point where the sequence entropy falls past the 5\% cutoff. We hypothesize that in these cases, the reduction in score occurs simultaneously with the degeneration of sequences beyond the natural range of IDR sequence entropies.

\newpage

\section{Random feature selection control}\label{app:control}

\begin{table}[H]\centering
\footnotesize
\setlength{\tabcolsep}{5pt}
\begin{tabular}{lccc@{\hskip 10pt}ccc}
\toprule
& \multicolumn{3}{c}{Target feature coverage $R_S$} & \multicolumn{3}{c}{On-target ProtGPS} \\
\cmidrule(lr){2-4}\cmidrule(lr){5-7}
\textbf{Target} & Base & random-30 & top-30 & Base & random-30 & top-30 \\
\midrule
Nucleolus & $0.03_{.00}$ & $0.59_{.04}$ & $0.99_{.00}$ & $0.14_{.01}$ & $0.17_{.28}$ & $0.99_{.00}$ \\
Chromosome & $0.04_{.00}$ & $0.55_{.06}$ & $0.99_{.00}$ & $0.26_{.01}$ & $0.22_{.26}$ & $0.71_{.03}$ \\
Speckle & $0.01_{.00}$ & $0.58_{.07}$ & $0.93_{.00}$ & $0.00_{.00}$ & $0.00_{.00}$ & $0.36_{.15}$ \\
\midrule
Stress gran. & $0.01_{.00}$ & $0.59_{.05}$ & $0.95_{.00}$ & $0.05_{.00}$ & $0.05_{.06}$ & $0.32_{.09}$ \\
P-body & $0.02_{.00}$ & $0.61_{.09}$ & $0.69_{.04}$ & $0.17_{.01}$ & $0.27_{.40}$ & $0.05_{.02}$ \\
\midrule
NPC & $0.00_{.00}$ & $0.56_{.04}$ & $0.94_{.02}$ & $0.08_{.01}$ & $0.15_{.23}$ & $0.84_{.11}$ \\
\midrule
Activation & $0.05_{.00}$ & $0.68_{.07}$ & $0.99_{.00}$ & -- & -- & -- \\
Repression & $0.11_{.00}$ & $0.62_{.03}$ & $0.76_{.00}$ & -- & -- & -- \\
\bottomrule
\end{tabular}
\caption{
RL-SAE on randomly chosen target features (random-30) compared to the base model and to RL-SAE on the enriched top-30. 
Subscripted values are standard deviations over three seeds (Appendix~\ref{app:stats}).}
\label{tab:control}
\end{table}

\paragraph{Random target feature sets}
As a control, we run RL-SAE on randomly chosen sets of SAE features (random-30). For each of the 8 sequence sets, we sample 30 features uniformly at random from all 16,384 IDiomSAE latents, excluding that set's enriched top-30. We draw three independent random sets per target and then run RL-SAE on them using the same hyperparameters as other runs. We generate and score 1,000 sequences per run. We find that while RL-SAE achieves high target feature coverage for the random features ($R_S \approx 0.55 \text{ to } 0.68$), the ProtGPS scores remain low, close to the base model's values, as compared to RL-SAE on the top-30 enriched features.

\newpage

\section{Oracle RL baseline (RL-GPS)}\label{app:rlgps}

\begin{table}[H]\centering
\footnotesize
\setlength{\tabcolsep}{6pt}
\begin{tabular}{lc}
\toprule
\textbf{Compartment} & RL-GPS ProtGPS \\
\midrule
Nucleolus & $1.00_{.00}$ \\
Chromosome & $0.99_{.00}$ \\
Speckle & $0.98_{.00}$ \\
Stress gran. & $0.99_{.00}$ \\
P-body & $1.00_{.00}$ \\
NPC & $0.99_{.00}$ \\
\bottomrule
\end{tabular}
\caption{ProtGPS-predicted compartment association scores (0 to 1) for the oracle RL baseline (RL-GPS).
Subscripted values are standard deviations over three training seeds (Appendix~\ref{app:stats}).}
\label{tab:rlgps}
\end{table}

\newpage

\section{Composition scramble controls}\label{app:scramble}

\begin{table}[H]\centering
\footnotesize
\setlength{\tabcolsep}{6pt}
\begin{tabular}{lcccc}
\toprule
& \multicolumn{2}{c}{RL-SAE top-30} & \multicolumn{2}{c}{RL-SAE private} \\
\cmidrule(lr){2-3}\cmidrule(lr){4-5}
\textbf{Compartment} & on-target & scramble & on-target & scramble \\
\midrule
Nucleolus & $0.99_{.00}$ & $0.50_{.03}$ & $0.94_{.07}$ & $0.39_{.02}$ \\
Chromosome & $0.71_{.03}$ & $0.38_{.04}$ & $0.67_{.08}$ & $0.44_{.02}$ \\
Speckle & $0.36_{.15}$ & $0.11_{.02}$ & $0.30_{.24}$ & $0.14_{.05}$ \\
Stress gran. & $0.32_{.09}$ & $0.68_{.04}$ & $0.51_{.36}$ & $0.56_{.07}$ \\
P-body & $0.05_{.02}$ & $0.33_{.03}$ & $0.32_{.10}$ & $0.17_{.01}$ \\
NPC & $0.84_{.11}$ & $0.50_{.10}$ & $0.90_{.05}$ & $0.57_{.06}$ \\
\bottomrule
\end{tabular}
\caption{ProtGPS scores of RL-SAE localization sequences and their composition-matched scrambles.
Subscripted values are standard deviations over three training seeds (Appendix~\ref{app:stats}).}
\label{tab:scram_loc}
\end{table}

\begin{table}[H]\centering
\footnotesize
\setlength{\tabcolsep}{6pt}
\begin{tabular}{lcc}
\toprule
\textbf{Readout} & RL-SAE & scramble \\
\midrule
PADDLE & $11.16_{.11}$ & $8.73_{.10}$ \\
$\psi$KxE & $5.63_{.60}$ & $0.49_{.08}$ \\
NDSM & $1.80_{.44}$ & $0.12_{.02}$ \\
\bottomrule
\end{tabular}
\caption{Transcriptional effector readouts of RL-SAE sequences and their composition-matched scrambles.
Subscripted values are standard deviations over three training seeds (Appendix~\ref{app:stats}).}
\label{tab:scram_eff}
\end{table}

\newpage

\section{Additional disorder predictions}\label{app:disorder}

\begin{table}[h]\centering
\footnotesize
\setlength{\tabcolsep}{4pt}
\begin{tabular}{lccccc}
\toprule
\textbf{Target} & Nat. & SFT & Steering ($\alpha$) & RL-SAE & RL-GPS \\
\midrule
& \multicolumn{5}{c}{Metapredict~V3} \\
\cmidrule(lr){2-6}
Nucleolus & $0.85_{.00}$ & $0.81_{.01}$ & $0.87_{.00}$ (0.35) & $0.93_{.00}$ & $0.95_{.00}$ \\
Chromosome & $0.86_{.00}$ & $0.81_{.01}$ & $0.88_{.00}$ (0.21) & $0.92_{.01}$ & $0.91_{.00}$ \\
Speckle & $0.83_{.01}$ & $0.82_{.00}$ & $0.89_{.00}$ (0.20) & $0.84_{.07}$ & $0.94_{.00}$ \\
Stress gran. & $0.86_{.00}$ & $0.80_{.01}$ & $0.86_{.01}$ (0.15) & $0.91_{.01}$ & $0.93_{.01}$ \\
P-body & $0.82_{.01}$ & $0.80_{.00}$ & $0.64_{.01}$ (0.00) & $0.92_{.01}$ & $0.91_{.01}$ \\
NPC & $0.85_{.01}$ & $0.84_{.00}$ & $0.81_{.00}$ (0.20) & $0.89_{.01}$ & $0.44_{.12}$ \\
Activation & $0.74_{.01}$ & $0.75_{.00}$ & $0.77_{.00}$ (0.43) & $0.83_{.01}$ & -- \\
Repression & $0.73_{.01}$ & $0.75_{.00}$ & $0.76_{.00}$ (0.35) & $0.90_{.01}$ & -- \\
\midrule
& \multicolumn{5}{c}{IUPred3} \\
\cmidrule(lr){2-6}
Nucleolus & $0.74_{.01}$ & $0.68_{.01}$ & $0.78_{.00}$ (0.35) & $0.87_{.01}$ & $0.91_{.02}$ \\
Chromosome & $0.72_{.01}$ & $0.68_{.01}$ & $0.81_{.01}$ (0.21) & $0.86_{.01}$ & $0.71_{.01}$ \\
Speckle & $0.67_{.01}$ & $0.65_{.00}$ & $0.79_{.01}$ (0.20) & $0.77_{.03}$ & $0.97_{.01}$ \\
Stress gran. & $0.71_{.01}$ & $0.67_{.01}$ & $0.70_{.01}$ (0.15) & $0.63_{.01}$ & $0.87_{.02}$ \\
P-body & $0.70_{.01}$ & $0.67_{.01}$ & $0.49_{.01}$ (0.00) & $0.89_{.01}$ & $0.82_{.02}$ \\
NPC & $0.64_{.02}$ & $0.63_{.00}$ & $0.54_{.00}$ (0.20) & $0.49_{.04}$ & $0.12_{.09}$ \\
Activation & $0.50_{.01}$ & $0.52_{.01}$ & $0.39_{.00}$ (0.43) & $0.48_{.03}$ & -- \\
Repression & $0.57_{.01}$ & $0.60_{.01}$ & $0.68_{.00}$ (0.35) & $0.78_{.05}$ & -- \\
\bottomrule
\end{tabular}
\caption{Additional disorder predictions for generated and natural sequences, using Metapredict V3 and IUPred3. Values range from 0 to 1, and higher values indicate a prediction of greater disorder. Subscripted values are standard deviations over three seeds, except for natural IDRs (Nat.), which report the standard error of the mean (Appendix~\ref{app:stats}).
}
\label{tab:si_disorder}
\end{table}

\paragraph{Predicted disorder of designed sequences}
In Table \ref{tab:si_disorder} we present Metapredict V3 and IUPred3 disorder predictions on IDR sequences generated from the various methods and sets. Both Metapredict and IUPred consider a sequence to be disordered if the score is $>0.5$. 

With one exception, we find that across all methods and sequence sets, Metapredict predicts the sequences to be disordered. The exception is RL-GPS generated NPC sequences, with a score of 0.44. As RL-GPS on NPC was able to achieve a high ProtGPS score, we posit that this is due to hacking of the ProtGPS reward. 

Sequences are also largely predicted to be disordered by IUPred, with most scores near or above the 0.5 threshold. Exceptions include sequences steered for activation, with a score of 0.39, and RL-GPS NPC sequences, with a score of 0.12. The high hydrophobic residue content in steered activation domain sequences may lead to a low IUPred score, despite their functional relevance to co-activator recruitment. Here, the extremely low NPC score again indicates the possibility of reward hacking.

\newpage

\section{Orthogonal localization prediction}\label{app:deeploc}

\begin{table}[H]\centering
\footnotesize
\setlength{\tabcolsep}{4pt}
\begin{tabular}{lccccccc}
\toprule
& & & & & \multicolumn{2}{c}{RL-SAE} & \\
\cmidrule(lr){6-7}
\textbf{Compartment} & Base & Nat. & SFT & Steering ($\alpha$) & top-30 & private-30 & RL-GPS \\
\midrule
& \multicolumn{7}{c}{$P(\text{Nucleus})$} \\
\cmidrule(lr){2-8}
Nucleolus & \exq{$0.43_{.01}$} & \exq{$0.66_{.01}$} & \exq{$0.59_{.01}$} & \exq{$0.47_{.01}$} (0.35) & \exq{$\mathbf{0.90}_{.01}$} & \exq{$\underline{0.87}_{.02}$} & \exq{$0.70_{.03}$} \\
Chromosome & \exq{$0.45_{.01}$} & \exq{$0.67_{.01}$} & \exq{$0.58_{.01}$} & \exq{$0.57_{.01}$} (0.21) & \exq{$\mathbf{0.91}_{.00}$} & \exq{$\underline{0.85}_{.01}$} & \exq{$0.75_{.02}$} \\
Speckle & \exq{$0.44_{.02}$} & \exq{$0.67_{.02}$} & \exq{$0.56_{.01}$} & \exq{$0.63_{.01}$} (0.20) & \exq{$0.80_{.04}$} & \exq{$\underline{0.82}_{.05}$} & \exq{$\mathbf{0.84}_{.03}$} \\
\midrule
Stress gran. & $0.43_{.02}$ & $0.61_{.01}$ & $0.53_{.00}$ & $0.54_{.01}$ (0.15) & $0.78_{.02}$ & $0.56_{.03}$ & $0.61_{.16}$ \\
P-body & $0.44_{.01}$ & $0.61_{.01}$ & $0.55_{.01}$ & $0.45_{.01}$ (0.00) & $0.81_{.05}$ & $0.42_{.03}$ & $0.49_{.09}$ \\
\midrule
NPC & \exq{$0.44_{.01}$} & \exq{$\underline{0.59}_{.02}$} & \exq{$0.58_{.01}$} & \exq{$0.33_{.01}$} (0.20) & \exq{$\mathbf{0.68}_{.04}$} & \exq{$\mathbf{0.68}_{.06}$} & \exq{$0.22_{.03}$} \\
\midrule
& \multicolumn{7}{c}{$P(\text{Cytoplasm})$} \\
\cmidrule(lr){2-8}
Nucleolus & $0.50_{.01}$ & $0.45_{.01}$ & $0.49_{.00}$ & $0.52_{.01}$ (0.35) & $0.18_{.02}$ & $0.30_{.03}$ & $0.43_{.05}$ \\
Chromosome & $0.50_{.01}$ & $0.48_{.01}$ & $0.51_{.00}$ & $0.38_{.01}$ (0.21) & $0.20_{.01}$ & $0.29_{.02}$ & $0.40_{.01}$ \\
Speckle & $0.50_{.00}$ & $0.46_{.01}$ & $0.50_{.01}$ & $0.35_{.01}$ (0.20) & $0.34_{.02}$ & $0.33_{.06}$ & $0.32_{.02}$ \\
\midrule
Stress gran. & \exq{$0.51_{.00}$} & \exq{$0.53_{.01}$} & \exq{$0.51_{.01}$} & \exq{$0.46_{.02}$} (0.15) & \exq{$0.56_{.01}$} & \exq{$\mathbf{0.67}_{.02}$} & \exq{$\underline{0.64}_{.08}$} \\
P-body & \exq{$0.51_{.01}$} & \exq{$0.50_{.01}$} & \exq{$0.52_{.01}$} & \exq{$0.50_{.00}$} (0.00) & \exq{$0.21_{.02}$} & \exq{$\mathbf{0.65}_{.03}$} & \exq{$\underline{0.55}_{.02}$} \\
\midrule
NPC & $0.50_{.01}$ & $0.48_{.02}$ & $0.48_{.00}$ & $0.40_{.01}$ (0.20) & $0.32_{.04}$ & $0.30_{.02}$ & $0.23_{.05}$ \\
\bottomrule
\end{tabular}
\caption{
Orthogonal DeepLoc localization predictions. Bold indicates the best score and underline the second best within each grey shaded row. The private-30 column targets the top 30 private features of a compartment. RL-GPS is included for comparison here because DeepLoc is an orthogonal predictor, unlike in the main text Table \ref{tab:localization}. The grey shaded rows indicate the biologically expected localization. Subscripted values are standard deviations over three seeds, except for natural IDRs (Nat.), which report the standard error of the mean (Appendix~\ref{app:stats}).
}
\label{tab:deeploc}
\end{table}

\newpage

\section{Compartment motif detection}\label{app:motifs}

\begin{table}[H]\centering
\footnotesize
\setlength{\tabcolsep}{1.7pt}
\begin{tabular}{llccccccc}
\toprule
& & & & & & \multicolumn{2}{c}{RL-SAE} & \\
\cmidrule(lr){7-8}
\textbf{Compartment} & \textbf{Motif} & Base & Nat. & SFT & Steering ($\alpha$) & top-30 & private-30 & RL-GPS \\
\midrule
Nucleolus & cNLS & $0.13_{.02}$ & $0.54_{.02}$ & $0.37_{.00}$ & $0.38_{.01}$ (0.35) & $\mathbf{1.00}_{.00}$ & $\underline{1.00}_{.00}$ & $1.00_{.00}$ \\
Chromosome & cNLS & $0.13_{.02}$ & $0.43_{.02}$ & $0.27_{.00}$ & $0.82_{.02}$ (0.21) & $\mathbf{1.00}_{.00}$ & $\underline{0.96}_{.03}$ & $0.64_{.14}$ \\
Speckle & RS repeat & $0.00_{.00}$ & $0.08_{.01}$ & $0.02_{.00}$ & $0.21_{.02}$ (0.20) & $\underline{0.97}_{.04}$ & $\mathbf{1.00}_{.01}$ & $0.27_{.26}$ \\
\midrule
Stress gran. & RG/RGG & $0.04_{.01}$ & $0.15_{.01}$ & $0.09_{.00}$ & $0.45_{.01}$ (0.15) & $\mathbf{1.00}_{.00}$ & $\underline{0.93}_{.04}$ & $0.66_{.52}$ \\
P-body & RG/RGG & $0.04_{.01}$ & $0.13_{.02}$ & $0.10_{.01}$ & $0.04_{.00}$ (0.00) & $\underline{0.34}_{.56}$ & $0.00_{.00}$ & $\mathbf{0.51}_{.02}$ \\
\midrule
NPC & GLFG /100 aa & $0.00_{.00}$ & $0.29_{.08}$ & $0.29_{.01}$ & $0.01_{.00}$ (0.20) & $\underline{1.62}_{.45}$ & $\mathbf{1.84}_{.05}$ & $0.00_{.00}$ \\
\bottomrule
\end{tabular}
\caption{
Fraction of sequences containing the specified biological motif (density per 100 aa for GLFG). Bold indicates the best score per row, underline the second best. The private-30 column targets the compartment's private top-30 features. RL-GPS is included for comparison here because the motifs are orthogonal to ProtGPS.
Motif regexes are in Appendix~\ref{app:nls}. Subscripted values are standard deviations over three seeds, except for natural IDRs (Nat.), which report the standard error of the mean (Appendix~\ref{app:stats}).}
\label{tab:orthogonal}
\end{table}

\newpage

\section{Motif regular expressions}\label{app:nls}

\begin{table}[H]\centering
\scriptsize
\renewcommand{\arraystretch}{1.25}
\begin{tabularx}{\linewidth}{>{\raggedright\arraybackslash}p{3.2cm} >{\raggedright\arraybackslash}X >{\raggedright\arraybackslash}p{3.0cm}}
\toprule
\textbf{Motif} & \textbf{Regex / rule} & \textbf{Compartment} \\
\midrule
RS/SR repeat & \texttt{(?:RS|SR)\{3,\}} & Nuclear speckle \\
RG/RGG cluster & $\geq\!2$ of \{\texttt{RGG}, \texttt{RG} not in an \texttt{RGG}\} within 30 residues & Stress granule, P-body \\
GLFG repeat & \texttt{(?=(GLFG))} (overlapping matches) & Nuclear pore complex \\
$\psi$KxE & \texttt{[VILMF]K.E} & Repression domain \\
NDSM & \texttt{[VILMF]K.E[DE]+} & Repression domain \\
\bottomrule
\end{tabularx}
\caption{Regular expressions used in Tables~\ref{tab:orthogonal} and~\ref{tab:effector}.}
\label{tab:si_motif_regex}
\end{table}

\begin{table}[h]\centering
\scriptsize
\renewcommand{\arraystretch}{1.25}
\begin{tabularx}{\linewidth}{>{\raggedright\arraybackslash}p{3.2cm} >{\raggedright\arraybackslash}X}
\toprule
\textbf{ELM class} & \textbf{Regex} \\
\midrule
\texttt{TRG\_NLS\_MonoCore\_2} & \texttt{[\^{}DE]((K[RK])|(RK))[KRP][KR][\^{}DE]} \\
\texttt{TRG\_NLS\_MonoExtN\_4} & \texttt{(([PKR].\{0,1\}[\^{}DE])|([PKR]))((K[RK])|(RK))(([\^{}DE][KR])|([KR][\^{}DE]))[\^{}DE]} \\
\texttt{TRG\_NLS\_MonoExtC\_3} & \texttt{[\^{}DE]((K[RK])|(RK))(([\^{}DE][KR])|([KR][\^{}DE]))(([PKR])|([\^{}DE][DE]))} \\
\texttt{TRG\_NLS\_Bipartite\_1} & \texttt{[KR][KR].\{7,15\}[\^{}DE]((K[RK])|(RK))(([\^{}DE][KR])|([KR][\^{}DE]))[\^{}DE]} \\
\bottomrule
\end{tabularx}
\caption{Eukaryotic Linear Motif classical-NLS (cNLS) regular expressions used in Table \ref{tab:orthogonal}.}
\label{tab:si_nls_regex}
\end{table}

\newpage

\section{Top-$N$ logo galleries}\label{app:galleries}

\begin{figure}[H]\centering
\includegraphics[width=\textwidth]{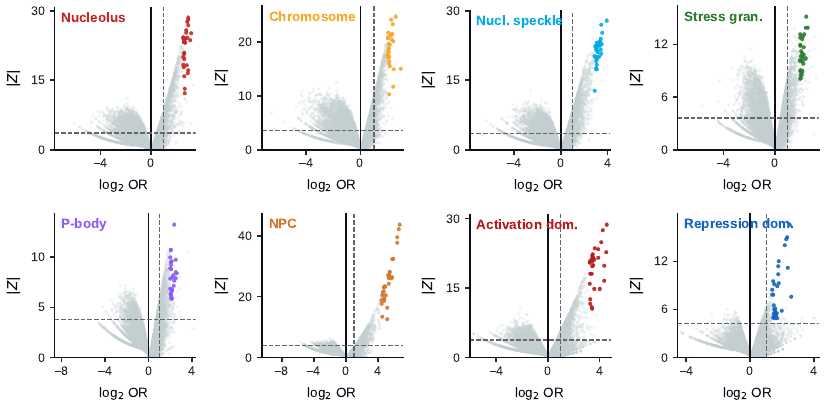}
\caption{
\textbf{Feature enrichment plots for all eight sequence sets.} 
Feature enrichment plots (hypergeometric $|Z|$ versus log$_2$ odds ratio (LOR)) for all 8 sequence sets. The top-30 features as ranked by LOR are the colored dots. The horizontal dashed line marks the FDR threshold, and the vertical dashed line marks the LOR $\ge$ 1 threshold. 
}\label{fig:si_volcano_grid}
\end{figure}

\paragraph{Logo plots}
Each of the following figures shows information content sequence logo plots of the top-30 enriched features for a given sequence set (row major). Each logo plot is centered on the feature's top-activating residue position (vertical dashed line) within a given sequence, with the orange color depicting the strength of the feature's activation, and is constructed by averaging across the top 80 sequences by feature activation strength from the corresponding positive sequence sets (pooled for localization sequences).

\begin{figure}[H]\centering
\includegraphics[width=\textwidth]{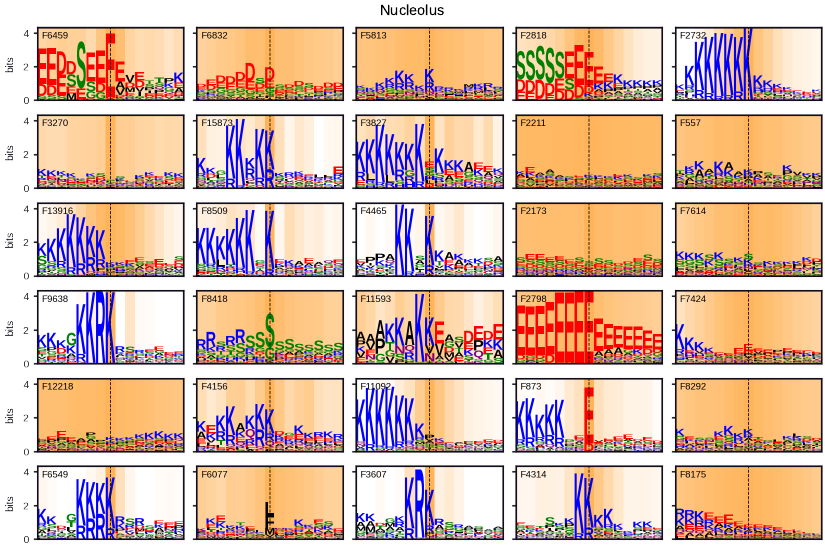}
\caption{\textbf{Nucleolus.} Logo plots of the top-30 enriched features.}\label{fig:gallery_nucleolus}
\end{figure}

\begin{figure}[H]\centering
\includegraphics[width=\textwidth]{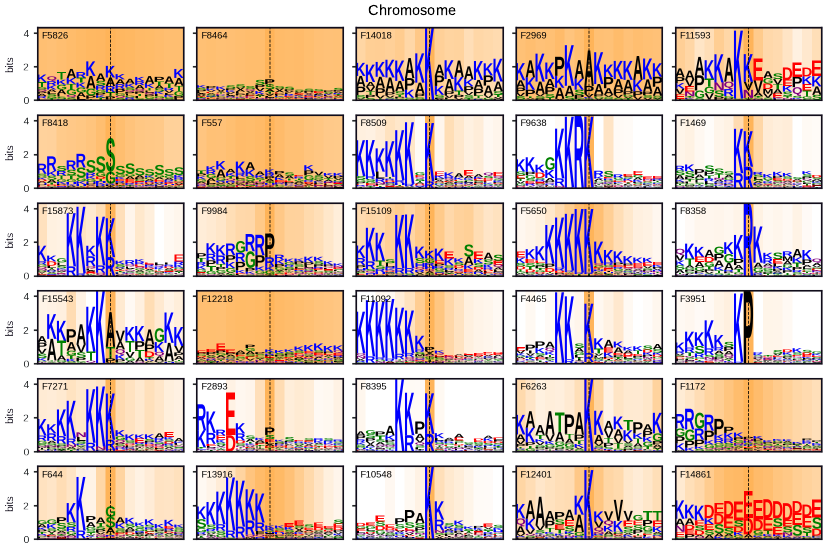}
\caption{\textbf{Chromosome.} Logo plots of the top-30 enriched features.}\label{fig:gallery_chromosome}
\end{figure}

\begin{figure}[H]\centering
\includegraphics[width=\textwidth]{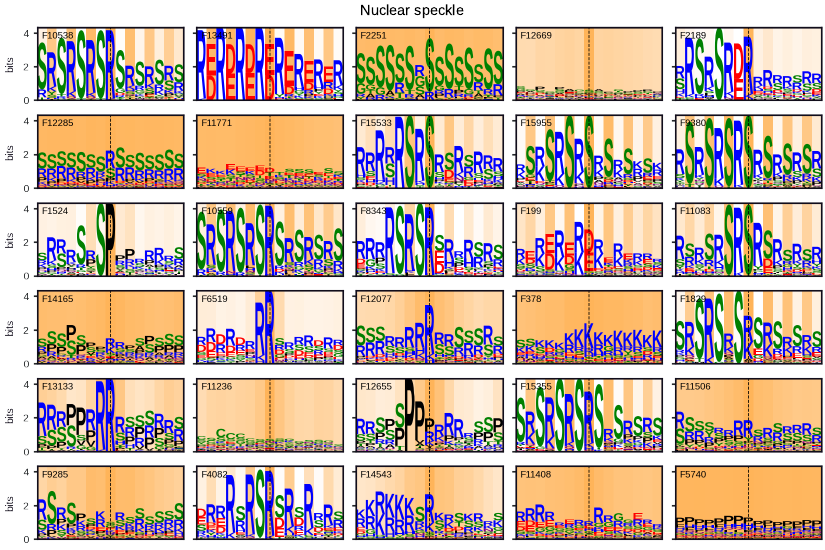}
\caption{\textbf{Nuclear speckle.} Logo plots of the top-30 enriched features.}\label{fig:gallery_nuclear_speckle}
\end{figure}

\begin{figure}[H]\centering
\includegraphics[width=\textwidth]{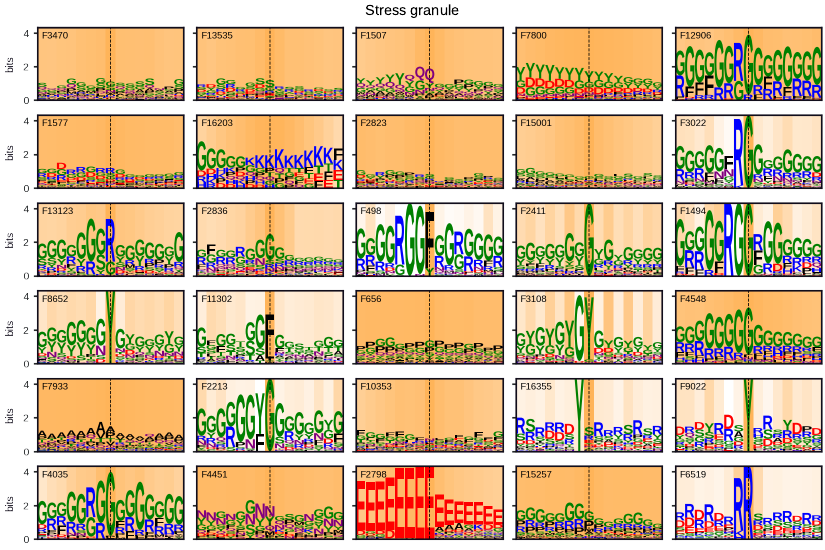}
\caption{\textbf{Stress granule.} Logo plots of the top-30 enriched features.}\label{fig:gallery_stress_granule}
\end{figure}

\begin{figure}[H]\centering
\includegraphics[width=\textwidth]{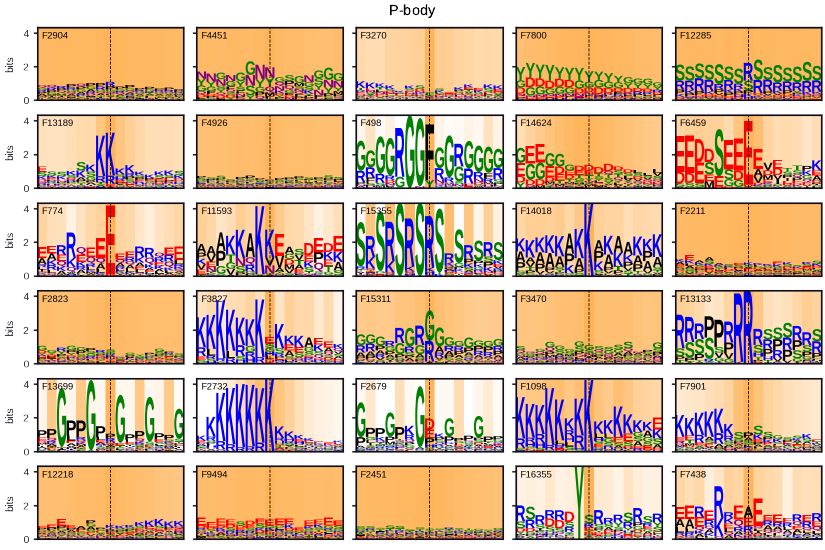}
\caption{\textbf{P-body.} Logo plots of the top-30 enriched features.}\label{fig:gallery_P-body}
\end{figure}

\begin{figure}[H]\centering
\includegraphics[width=\textwidth]{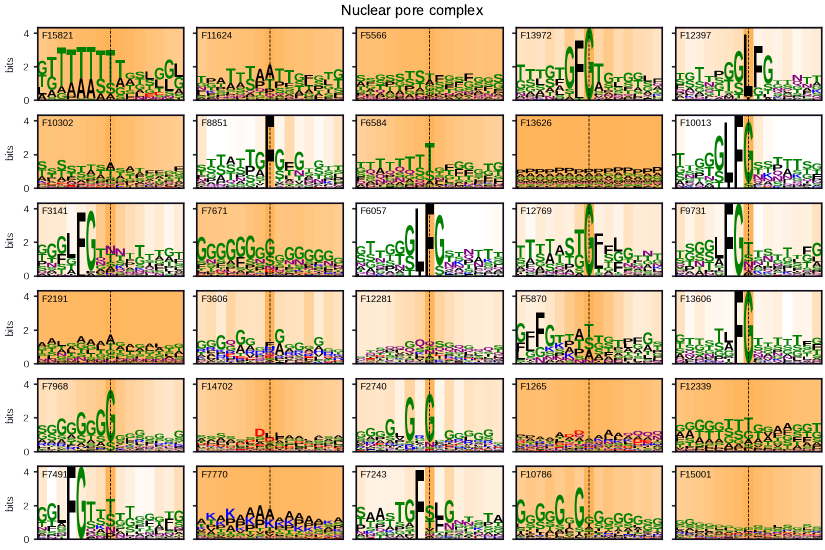}
\caption{\textbf{Nuclear pore complex.} Logo plots of the top-30 enriched features.}\label{fig:gallery_nuclear_pore_complex}
\end{figure}

\begin{figure}[H]\centering
\includegraphics[width=\textwidth]{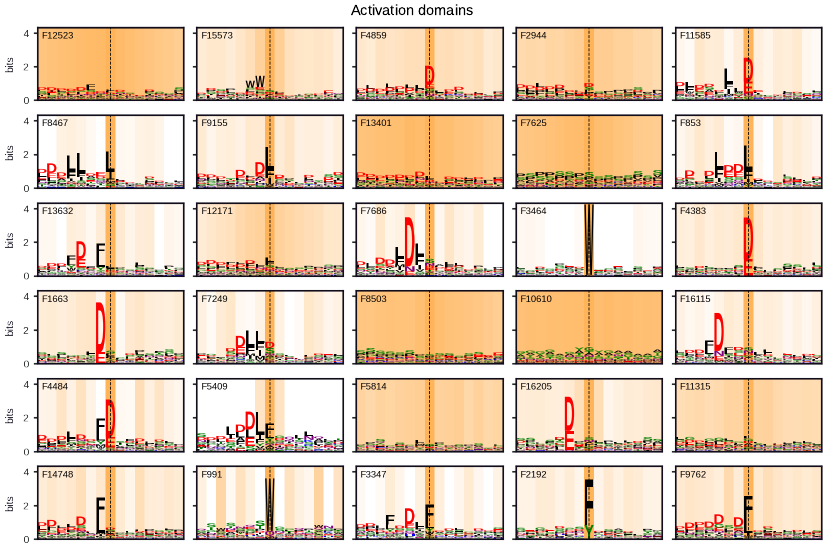}
\caption{\textbf{Activation domains.} Logo plots of the top-30 enriched features.}\label{fig:gallery_ad}
\end{figure}

\begin{figure}[H]\centering
\includegraphics[width=\textwidth]{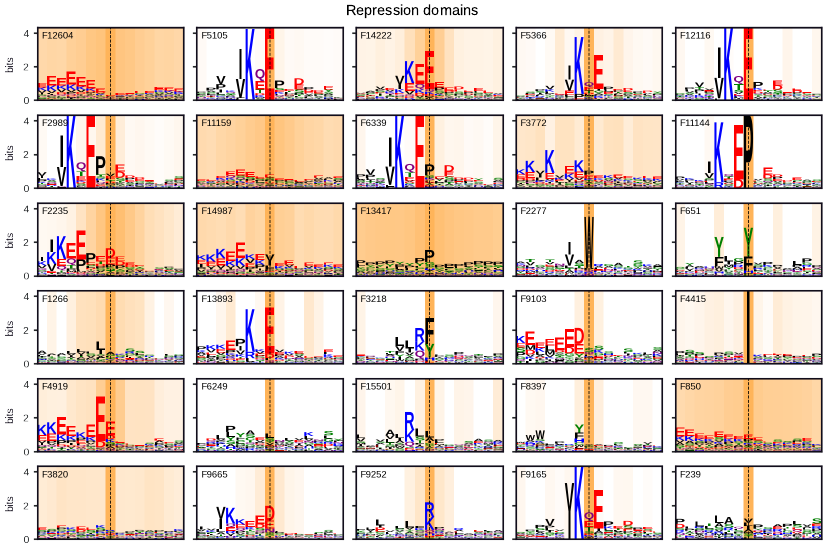}
\caption{\textbf{Repression domains.} Logo plots of the top-30 enriched features.}\label{fig:gallery_rd}
\end{figure}

\newpage

\section{Example sequences}\label{app:sequences}

Here we show five randomly chosen example sequences per condition.

\input{example_sequences}

\end{document}

%% file: math_commands.tex
\usepackage{amsmath,amsfonts,bm}

\def\eqref#1{equation~\ref{#1}}

\def\1{\bm{1}}

\def\vb{{\bm{b}}}

\def\vd{{\bm{d}}}

\def\vx{{\bm{x}}}

\def\vz{{\bm{z}}}

\def\mW{{\bm{W}}}

\DeclareMathAlphabet{\mathsfit}{\encodingdefault}{\sfdefault}{m}{sl}
\SetMathAlphabet{\mathsfit}{bold}{\encodingdefault}{\sfdefault}{bx}{n}

\newcommand{\E}{\mathbb{E}}

\newcommand{\KL}{D_{\mathrm{KL}}}

%% file: example_sequences.tex
\subsection*{Nucleolus}
\begin{genseq}{RL-SAE, Nucleolus ($N{=}30$)}
\seqsplit{KKKRKKGTKRLKDAEKTAQKKKKMKQKGPQGAEDSSTSEEESEEEGQKDEDEGDDFFVGHEKTEPKTDKMPKHKKHKNQMQTTLHANGDSDSEDEEDDIVLQVTSGDIA}\\[6pt]
\seqsplit{KKKRKKEKKGGRETKTAKQRKKRKQADEKGERDSSSEEEEIEEEGPGPVGNDNDDFFQNKEEASKQQRPKKQAAPHKKAKMQHHSTASNPLTDSDKSEDEDEDEFDDDIVL}\\[6pt]
\seqsplit{KKKRKKGTKAADQLNEKKVSKKRKEKQKAAPMTEDDSQKSSSEESEEEKEDDGEDKVKAEDDGDDFFVGTNHKEQPTLKKHKTGKKHTVQNPNHNSDSDESDEEDEIVLML}\\[6pt]
\seqsplit{EKVKKKRKKGSKAAVEKASKRRKQSKQTKEPVESESSSEEEEEEEEDAGDSDDFFVGEDTSRDAPMKLKKNKTKVQPHHKGAKPQQQNHPQPVAG}\\[6pt]
\seqsplit{AKPKNKKRKKEPQGKKGAKKEKQAKAHGAGSEEESGSSSEEEQEQDTGGDDGDDFFTDKRSDQLPQQPTKKMAKKHKSATKDHNQQTAPHRIPDSSDEEEDDEDDILL}
\end{genseq}

\begin{genseq}{RL-GPS, Nucleolus}
\seqsplit{AATDDEEDRTAKGHPTGVQAQKSDQKGPKKTLRKPKKKGRKSRKKKRQKTEDPDLVLDREPTSTSTNEEESDREEDQKDTSEENETDQHRSQITKQNLN}\\[6pt]
\seqsplit{VKRESKAAFSTTATTQKEGPPENRKKTAKVKKVKKKRRKESKKQEGKVEPSDVASKDDIDDSQTVKNARSNQDPDDDAESNVEAPSMRNRSRDGVEKQTHPT}\\[6pt]
\seqsplit{LQQQGPGPGSAPFRPPDPKSTTPKEKGKPRRKAAKKRQKRRRKKEKRKKRALEAENSDEAGGLAVSVESESQHEVEGSSSLDSAEEADTEKTPQRRPGQPPVN}\\[6pt]
\seqsplit{KANAAGTESKSVSENKTRQKRDRKEKKRKRKNAEKKQPQDTTTQDANESSETSEDEEEEVERPNDDAPQQSRPEELPNAPQDDRVHEFQDTPPGDKRVRTGQ}\\[6pt]
\seqsplit{DKSGKSTAEVNRSSSKKRKKHKKTKKKATRKRRELKEDQVEDQSTNTREPDQRSEDADGDDSEPETSENPDTTADEQRGTEHPKSQNSHEDRFKGHPHNNLL}
\end{genseq}

\begin{genseq}{Activation steering ($N{=}30$, $\alpha{=}0.35$)}
\seqsplit{VENTKAKPVNKRKTEGDEEEGETQASKASAKKAKVEPAKEGSSKAAIKEEPAKEVKSEVKEEKKEEPVKKAASKEKKVEE}\\[6pt]
\seqsplit{TRLPVTSGSVSPQKTPLVSKRETGEDNEKGIPSLEEVAQRAQIADKKKTAKASKKKKDAEKKKEAADMSGEKLQRKVAREEEAKKQAAEKAAKAKAEK}\\[6pt]
\seqsplit{TARPPEEGEAPAGGEAPAGEGKAEGGVPSSGTSADAAEGSAKAAKGEPIGKKRAAAADEGGAAKKARKEEAGGDDKKSAAKAESAGSGKPAAGGGQGKARAGGGKAAAAGKKRKAEE}\\[6pt]
\seqsplit{FSPTRDEAAGESSQKPEKRKKLREAKEDGDPGLEQGDENDQEQEESDKELAKQSDDVEEAEENGEVEDEEHREQEGKEKPKKESSKSEEAQENGKEAEKAKKDENN}\\[6pt]
\seqsplit{TLTEVHTEKFQKDHKQQKVEQSTDQQHVSDTQKAAQQNVKATEEVEPRKAAPQAESDSVVESEKEKTPPAANTKPKKKTKK}
\end{genseq}

\begin{genseq}{Supervised fine-tuning}
\seqsplit{VTRPAGTSEGTILHQRALSSPKPVHRHGLSSLPQVVFTTAESTSKWEDVPLDRGSYLDDLNMEALSIKQDISGNDVSASLLPGERR}\\[6pt]
\seqsplit{LTYTVGLRRGFASSPQSWDRDDAAIRRAGSDIYSRISRGMPLPDKGRDEAPGWK}\\[6pt]
\seqsplit{MAAEKQVPGGGGGGGSGGGGGSGGGGSGGGRGAGGEENKENERPSAGSKANKEFGDSLSLEIL}\\[6pt]
\seqsplit{MAAPLMDQRARGQGDVHRGKLKEPLNLLQGRHRGQRRNRGDGGGRQPLPGLRRALPQRGACPLPALPDRGVNMAQRPMEQPMMMMM}\\[6pt]
\seqsplit{YKSLRRKPLIPLVSEPTFKDGFSECDSKDSELGKILQRDDCPPEGMMGCSASGELGSPPESCYHSPSNGNKPRDSASMDGPKYRGSGDSGTESDGGTKNKVATWPRQQPKHDPTQPSSDLGSGQP}
\end{genseq}

\begin{genseq}{Natural nucleolus IDRs}
\seqsplit{MTRILTAFKVVRTLKTGFGFTNVTAHQKWKFSRPGIRLLSVKAQTAHIV}\\[6pt]
\seqsplit{MNRSRAIVQRGRVLPPPAPLDTTNLAGRRTLQGRAKMASV}\\[6pt]
\seqsplit{MPTVISASVAPRTAAEPRSPGPVPHPAQSKATEAGGGNPSGIYS}\\[6pt]
\seqsplit{MEPQPGGARSCRRGAPGGACELGPAAEAAPMSLAIHSTTGTRYDLAVPPDETVEGLRKRLSQRLKVPKERLALLHKDTRLSSGKLQEFGVGDGSKLTLVPTVEAGLMSQASRPEQS}\\[6pt]
\seqsplit{MPGANYRAGAGAGAGARRPRGARDREEDGGGLEPAAVARDLLRGTSNMSFEELLELQSQVGTKTYKQLVAGNSPKKQASRPPIQNACVADKHRPL}
\end{genseq}

\newpage
\subsection*{Chromosome}
\begin{genseq}{RL-SAE, Chromosome ($N{=}30$)}
\seqsplit{KSRDEDKPQKISKKKSKTPNKKAKKAPRKPPGKRARKTLKKTAEASENMELDCGNTQDEFEKISAEDGPVKEEHPVKKRKRPPKLNSTPAKVHQTSDNKDVAL}\\[6pt]
\seqsplit{EDSSPKKTKKKKKSPVKKAKGPRKAGRPKKATRKTTEKLISTLESEMEENGLREESMNNEDD}\\[6pt]
\seqsplit{EVENDKSPKKKKKSKKPPSKKVKGRPAKTRAKRAARKTQELIHEENSLNSEDQEENLDQNDDVNEPPAKRKRKAQELSKSHSTKSDTPIKKTKGAPCTRCN}\\[6pt]
\seqsplit{DDPRTKETVVNKKKKNKKPPAKKAKSGRPRKAPRKTARKMQDTSDLDCEEQSGEESASEVPAQKRRKRKPQESADSSVVKQKRQPTSKVSQKCFEVNTD}\\[6pt]
\seqsplit{SKSGEKNDATIEKTKKKKKVPSRKTKKPPVKKIARKTARKKMASKPSDNEDSENEVVPSIEENDLSVPQKKRKRKPPQPGDVVECSLTQNNKRKRGVAG}
\end{genseq}

\begin{genseq}{RL-GPS, Chromosome}
\seqsplit{KSSSSGSPESTTPQRRRSQKSSTCLTQKKVNLHNRSAVESSRICNKRSQRSSCAEDFEDLSGSASGTAVSDEDSTDKSDDFESNGQKASRSLKEGSLASNKRHS}\\[6pt]
\seqsplit{EQRISELTLNSKPSKGNSCKASAGKKLYSCTVSSSRRSTNLTGSKQSKHARESP}\\[6pt]
\seqsplit{DKLQKQQADARSASTCKEQSCSVTPLKKSQSLRKSDSPEIQNNSCCLDSDFNSSPTKRQKLICSVENHPSVELVTGSELKKTECLDSDKRNGQPRSVT}\\[6pt]
\seqsplit{SKDTGKKASLCTSSILSEAKSLQGSRASSLTNGKSQPSEHNDSKSLTVTPQKTTSDKQTYCDSENTLIQRSSKKDPSTQPHPSNKEDPASDEETDASPNKRKRKPRQ}\\[6pt]
\seqsplit{RDQGSSCRKSKRTSSTCCKSRKSIQNSQEGFPLKKPTGKKKPCRKSSVDYTEESELEAAETVEGADQLSISKASRRAKSPQDTSGIENTEETSDS}
\end{genseq}

\begin{genseq}{Activation steering ($N{=}30$, $\alpha{=}0.21$)}
\seqsplit{VAATAERLSKTPKKDRDMKRIIEAVIKKKKEKKKAAQKAEYEKIKKRAAALKKKKKKIGKKKR}\\[6pt]
\seqsplit{RPSVPADHGALATALERSRAAVRPQPMKRVSRAANAASVAARALVKAADLMSVRRGVAAVKASAAPKATATKAPAVKATAAKATAAKPTAKTKSPAKRKATPATSRTRRRKSTDDAPAASPRKRAAKGPAKAARKRASKR}\\[6pt]
\seqsplit{MSSQPLSRRRPHWLRPLLRLAVAVVVLFTVAWLKGPDVLSPNDTALKAVASTTSSEKPLSTPVLMEKGPGKKSKDRTSSQANSGERSTSTSKRWTQTEDKTASEKGEKVERPEKKKKKVREKKKA}\\[6pt]
\seqsplit{LSVSGTMTGSSASAAGSEAPAPEQKEPKKVSGSSKKKKPAASKKKAAGSKEAKKRRKTRKKSAKKTKKATAKKAAKAAKTVKKKTKASKSKKASKAYKSKKKTKKTAKKAKKAKKSSKKKVTKSKKKTKASKSKKKAK}\\[6pt]
\seqsplit{MRRGLKRNWDGYINKRHREKRRLDARALAGGNIDLKGLKSKAGKAGKKRAAEKKPAAKGKAGKPPAKKPAKKPAAKKPAKKPAAKKPAAKKPAAKKAAKKPAKKPAAKKPAAKKPAAKKAAKKPAAKKTKKK}
\end{genseq}

\begin{genseq}{Supervised fine-tuning}
\seqsplit{MVKQSQHCPGTLGQESGLENHRDRAATKKTRGNSFERHSPLSNSPHKEPFLKHHHTGRKGERRSPKRRPGAPPVKNEPLTPVITGNHGKSV}\\[6pt]
\seqsplit{QPLNPKPKPKKEKTEQDKRRIEAEINTLTVESPWRAPPDLSPRPHPQNMPPKARLMRWGEFVEDPLLKQEVLVPGKKKTPP}\\[6pt]
\seqsplit{CSPAPSGAALAAGDRGPDPAPAVSDTGSQGGQPGPAAKTASPSASPSPKAPAKGA}\\[6pt]
\seqsplit{MSAQKDCEFWDQHEAKGTPQARRDDSYSPAEDASSKLVRKQTGNEEAGSYKAHRPLGKQHKLVTKPLG}\\[6pt]
\seqsplit{KPHSPKHARRTRSRHDQEPEAMVADSGSEDQHSVASDANRGREILAFSRPAAAAAPPPPQSQQQKSSRSGPNHHHNSGGGGSSRRKPTNLSPATQQAILSMLEDNQDPVIKGHGKEKNCA}
\end{genseq}

\begin{genseq}{Natural chromosome IDRs}
\seqsplit{ETDIFKKKKRKGQEDRKSKKKSAPSILSNGRIGQVGKFKNGTLILSPVDIKKINSSRVAK}\\[6pt]
\seqsplit{ESIYKTEPRPEAKTMPNSSTPTSQINTIDSRTLSDVSNISNINASCVSESICNSLNSKLEPTLENL}\\[6pt]
\seqsplit{TVAREATELNEDGSQVKRTGGAEDGAPAPFI}\\[6pt]
\seqsplit{MPRKGTQPSTARRREEGPPPPSPDGASSDAEPEPPSGRTESPATAAETASEELDNRSLEEILNSIPPPPPPAMTNE}\\[6pt]
\seqsplit{PSLEEVEGHVADGSATEMGTTKKAVKPKKKHLHRTIEQNINNLNVSEADRKCE}
\end{genseq}

\newpage
\subsection*{Nuclear speckle}
\begin{genseq}{RL-SAE, Nuclear speckle ($N{=}30$)}
\seqsplit{LSRRSTASRSRSRSRSRSRSYSRSRSPRRDRDKSPAKDKSRSKSPAKSSSRTPERDSHERDLDKHGHKDGDRNHCNCEHERNHAPAVEQNMCTNIIQLTTF}\\[6pt]
\seqsplit{DTARSRSRSVSRSRSGSRSRSYSRSRSPRRDRKHEAASRSRDRSPSKSPAKSLNQSKEHSREKDRDKNCDCGHERNNCHGAECNADADYVLDKQMVFATFPP}\\[6pt]
\seqsplit{IQTRSRSRSRSRSGSRSRSRSYSRSRSPRRERDRSPGKERSRSKSPANAAASHSRDKDRDKHREKDHHEHKNEWCEHGCNCGNEKPAQIANPNMGMPFI}\\[6pt]
\seqsplit{RTRTKNSHDRARSRSSSRSKSRSRSRSYSRSRSPRRERRDKDKSQTRSPSQDKGRSRSRDHEHDYQKDGKDCNCRETDKDCSCGTDQNMIINTGPFLIGEFP}\\[6pt]
\seqsplit{ARRSRSRSRQRDRSRSRSRSRSYSRSRSRSYSRSRSPRRSKDKDRERGASPSRGKEKRSVTRDSSRENHGKDRDHECADCENHSCDHNVDIDTCMMQFFTMQTF}
\end{genseq}

\begin{genseq}{RL-GPS, Nuclear speckle}
\seqsplit{ADPKHAKKRKPTDHRTYSRHSRTSKHRSSSRSRSRSGSHRRKKSRDRDKRRAEERKGSPPRTPATHVHQHDNHYGGAAGVKESANMAEMGDHVGISKIPQH}\\[6pt]
\seqsplit{SHHNHKKQQRDSRHPTSHDRRHNRSRSHSRSRSRPRDSRDRKSKKHTSSRRRSRSHDRDDGKKSKDPSANAPVDIVPTEMQPEDEMRLHHHDNIDLPPEIQ}\\[6pt]
\seqsplit{RAGHHKPPSSHQQRHRPRHYPQAHRKSTAASDRSRSESQRSRRHRSPDRGHHREKRKRTPQREGTQEKENKPHKEPANSMGNPQPNQHDQPNQDDHHDMAEGI}\\[6pt]
\seqsplit{DHHHHHPNRTPSTSRSRRHSNRHRSAKHSRSHRAHDKGRHPPNRKPRDYPPVEKGHRREAVAEAHGPGAIGNSGHNPLDSQHQDMHEMKRDHHQGKMESHDLP}\\[6pt]
\seqsplit{GAAHRHQHIKEKRKTIHPAHAHRKSASRRSTSKHDRHRSRRDRTKKSSSRREEPRKQRSAARDATQDRRTPAPIQEQTTKAPEMNPHDHNEPSRNVHEDLDHHHGQ}
\end{genseq}

\begin{genseq}{Activation steering ($N{=}30$, $\alpha{=}0.20$)}
\seqsplit{LMGSASSRQGHSEDSAFAAERQGTRNRKRGRRDRDHCSDNERRVNERRSDDYGSRRRSRRRCRSRRENSSRSRSRRSSRSRSRGRRSRSRDRRRGRRSRRTHSRSRRSRSRSRRRSRRSRRSRSRRRGSRSRSRRKR}\\[6pt]
\seqsplit{TGDEPRDTDANLALSPQPAAANTEDGPWVRMELHPYMARNYSQCIFRDDQGAARRRR}\\[6pt]
\seqsplit{GKAKDDPETTPAPKQSKPRPPKGKRNPRSRRKPKSAKTKSSPSRKRKSKRR}\\[6pt]
\seqsplit{SPQRSTETLKQTEAPYSSKRQRNDSTASDSSDTPIKRLKGNDNDEESPDAS}\\[6pt]
\seqsplit{INPAGGSSRGGSSRRAAVESLILDSRIMGGHGKHHRSGSRGHRSRSRGSRSGRDRSRDRDRDRDRERDRVDRTRSRSRRSRSR}
\end{genseq}

\begin{genseq}{Supervised fine-tuning}
\seqsplit{MADGVDHIDIYADVGEEFNQEAEYGGHDQIDLYDDVISPSANNGDAPEDRDYMDTLPPTVGDDV}\\[6pt]
\seqsplit{MPVLSARRRELRDPLRSLSTPVVATCLALGIVLLAGCTAAGSDTAAGDAQARAGGAAADTPEAP}\\[6pt]
\seqsplit{MQSNKTFNLEKQNHTPRKHHQHHHQQQHHQQQQQQPPPPPIPANGQQASSQNEGLTIDLKN}\\[6pt]
\seqsplit{MASGPHSTATAAAAASSAAPSAGGSSSGTTTTTTTTTGGILIGDRLYSEVSLTIDHSLIPEERLSPTPSMQDGLD}\\[6pt]
\seqsplit{MLGKDVLNPDFQKCGRLPATEDSSLHQEDSNPLTLRTSRPTGCRYSSQSHHGIYGRGTRGIKGKEPQGFYLSFREN}
\end{genseq}

\begin{genseq}{Natural nuclear speckle IDRs}
\seqsplit{LLEGAPQEEAEAAPEGNVGAEVVGEPRANIPCK}\\[6pt]
\seqsplit{RQEEERRRREEEMRRQQEEMMRRQQEGFKGTFPDAREQEIRMGQMAMGGAMGINNRGAMPPAPVPAGTPAPPGPATMMPDGTLGLTPPTTERFGQAATMEGIGAIGGTPPAFNRAAPGAEFAPNKRRRY}\\[6pt]
\seqsplit{LSTIKSSVFSLDGGSSPVEPDLAVAGIHSLPSTSVTPHSPSSPVGSVLLQDTKPTFEMQQPSPPIPPVHPD}\\[6pt]
\seqsplit{YGRRRKKHRSRSRSRERRSRSRDRGRGGGGGGGGGGGGRERDRRRSRDRERSGRF}\\[6pt]
\seqsplit{FPRSGRGTGRGGGGGGGGGAPRGRYGPPSRRSENR}
\end{genseq}

\newpage
\subsection*{Stress granule}
\begin{genseq}{RL-SAE, Stress granule ($N{=}30$)}
\seqsplit{SQSHSVTGRGGFGRGRGFDSGYGGRGTSRGGAGYGYSSRGDGSYGSRRNDGYGSYGNQSYGYNSQSSRRFGNAQGGPQSMHAMRIPMPSLHIPIPPVAIPNEIAVPVLE}\\[6pt]
\seqsplit{QSSTATSSMRGSGGRGFRGRGFDSGYGGRGGSSRGGGGYGYSSRRFDGGYGNGGYGNRDAYSNQQQRFGQSQQNRQARGMAMQIAPPQVRMPIPVAAAVAVPVLEVECNV}\\[6pt]
\seqsplit{ISSASTTSTMAAGRIGSPRMGGRGGFRGGFDSGYGSRGGSGYGYSSRRPDGYGSYGNRRGDGYANGYGNDRFNNQQNFGSRSQNMRSMVGVPMQNIHVPPMGIAAPAPVITTLPTPVE}\\[6pt]
\seqsplit{NPSASSTMSTADLGPRGGLRGGRGFGRGFDSGYGSRGGGGYGGYSGGRRFDNGYGNGHSAYGNAERRFGNGYSQQQQGGISFNHAQPIRMPPRVPQTPAGVPIVVNPPTRVIEQ}\\[6pt]
\seqsplit{VSPSASTQSMTGTGRGGFRGRGFDSGYGGRGGSGYGYSSRRAGGDFSGGYSSRRSDGYGNYNQNRHNRDNYGAQQQQVRMPRSTAMGGSMPVLMMPPVIAPALKME}
\end{genseq}

\begin{genseq}{RL-GPS, Stress granule}
\seqsplit{DQRTPGQEVAQVRAPGRDEADVKGRDARRNVDASEKRGALKGDKSYEKYEREGAKQGRNNGNGTKKGGEDTDGEDSEQVAGNSNGMNGNGNGNLDGGQTGNMVGN}\\[6pt]
\seqsplit{TAADEPAAQQTDAGRQLMGQHRGQLSNQRGPFAQANSGNAAPIGRVDNNRRGNGRGRGRRDNGNRGFRNGMPRMEGISTPVSNQPDIGAGGYNIGSPTGNLLQSS}\\[6pt]
\seqsplit{SLVSPAATTGKPDVVAEQPGSQNPRAINPPSGLRGDQQPANGKLNSEGNRGNGNRNRGGRNGWRRGGGWNNRNTYGGAEQMNNQTTGNNMGPTRVDIGPDTGLGA}\\[6pt]
\seqsplit{TSQAGNGVNPAMMRGQIPPNRIANTSQGNSRSDNRRNEGRDGLGRGNFRRGNNRGNGWRDSQGENIPVPSFAPQNISPSLSQRNATFGQISAPTSSTNFN}\\[6pt]
\seqsplit{ENQQSAPTQAQPNSVNGVSQRDAPKTQDPAAVSQNTRFGDGLGWDMNQSQNSSNGYGRNRNNRGGRGNRGNNAGFNGPNRMSGNNTVTMVSQITGTPSSSYN}
\end{genseq}

\begin{genseq}{Activation steering ($N{=}30$, $\alpha{=}0.15$)}
\seqsplit{LDYSTYIPKYNTTNQNWLEKRRAEQTNSNEFQIRRKLAYINMFDNNGPGSFDILSNTPNN}\\[6pt]
\seqsplit{MKSRNRLEINLLTGQNSSVFIQRVQRAVANASARGEYSDDAELGRFLLDGKLSANGDGAG}\\[6pt]
\seqsplit{QPNCRVPVPVPVHPPPPRGGQPGGPPPCPGPCMGPPGPVPVPVPVPCAVPRRGPGPGPVPAPGPRKS}\\[6pt]
\seqsplit{EVARYSPGEQSADSVKSFENRQPAVNEAVPSHTSASLNPALNSAFKRPAGATSSQSFRVPSMGDIGRIHPGF}\\[6pt]
\seqsplit{GKAKDDPETTPAPEQSEPRPPRGPRNPRGRGGPRGGFGGSGPSRGGGGGGGGGGMRSGPGGFFGFGGGGGRGYGRGGGYGRGGGGYGRGGGRGGGGGGGGRW}
\end{genseq}

\begin{genseq}{Supervised fine-tuning}
\seqsplit{MQQMGQGGMPGMGGGMPGMGGGPGRGAKGKQQPRKKKKKKGFGNLPKNFENSGPQGGQQSGGLGEDLPPGLGGGEKKQEGGLGGLPPGMELPKGQ}\\[6pt]
\seqsplit{DDANLGNDPPVKRKSQSWTSKILRKKNGVISKKIFIEQEEETPLPEVNQL}\\[6pt]
\seqsplit{MATDSWALAVDEQEAAVKSMTNLQIKEEKVKADTNGIIKTSTTAEKTDEEEKEDRAAQS}\\[6pt]
\seqsplit{QLIPVTPRTTRVNLIEVSGKDNEEALLNLEAPSDEEGSVPASSFSQVNCVDVKEEWQELSELSKACR}\\[6pt]
\seqsplit{HRHNVSSFQNGWAVHSTDLYDLCRELHGIVRVYAVQTGQKPPNALLSQKLSPP}
\end{genseq}

\begin{genseq}{Natural stress granule IDRs}
\seqsplit{QRKLRESQEGREVQRQEAGELRRSLGEGAKEREALRRSNEELRSAVKKAESERIS}\\[6pt]
\seqsplit{KVKMKQPKGSLQAHDTSSLPTVIMRNKPSQPKERPRSAVLLVDETATTPIFANRRSQQSVSLSKSVSIQNITGVGNDENMSNTWKFLSHSTDSLNKISKVNESTESLTDEGVGTDMNEGQLLGDFEIESKQL}\\[6pt]
\seqsplit{MGLCLAGRRRRSVQWCAVSQPEATKCFQWQRNMRKVRGPPVSCIKRDSPIQCIQAIAENR}\\[6pt]
\seqsplit{MPLGLLWLGLALLGALHAQAQDSTSDLIPAPPLSK}\\[6pt]
\seqsplit{MACRPRSPPRHQSRCDGDASPPSPARWSLGRKRRADGRRWRPEDAEEAEHRGAERRPESFTTPEGPKPRSRCSDWASAVEEDEMRTRVNKEMARYKRKLLINDFGRERKSSSGSSDSKESMSTVPADFETDESV}
\end{genseq}

\newpage
\subsection*{P-body}
\begin{genseq}{RL-SAE, P-body ($N{=}30$)}
\seqsplit{NADEDKTKKEKKKKRKKRKDKEESSSDEEDEEIDEDLPQPHQVKKSKKKKRKHKEGADDSYDDDYATPKVNHAGSKSKGSNIPLPGQGAMPPPGQVSKNTPRHASIEGP}\\[6pt]
\seqsplit{DADEKKKKKRKTRDDDESSSEEEEVDEELPQPHQIKKKKKKRKHKAYDDDDYDYDDNYDDGSKKNKQQSETPLPGQGSYVPPHTKKIPAGDGSPVTPGAPTGHGNT}\\[6pt]
\seqsplit{DADEKKKKKRKKRKDKEQDESSESEEEEIDEDLPQPHQLPHPKKKKKKRKREDGEDSDNESDSDSSDLDNQEEDNLNPGSIPLPGQGKMKVPSGHPHNQDPPPSDSGGP}\\[6pt]
\seqsplit{DADEKKKKKRKKRKDKDEEESSSDEEEDEIDESDLPQPHQMKKKKKKKHKQDDDYDDDEYDERDYRGKTSGGVSLPGQGNRLPGPPKPGTHLSATGNRTNSTASEPQGAV}\\[6pt]
\seqsplit{DADEKKKKKRKKRKDKEDEESSGEKDEEEEIDEDLPQPHQIKKKHKKKKHKDAQEEISDDYDDDYDDQQTTGGKNKQEDLPLPGQGGLPVPGQGAQVPTPDLPNKNLS}
\end{genseq}

\begin{genseq}{RL-GPS, P-body}
\seqsplit{TAGNAGKTDSPNPASKVDDNGKGIIGATLEKPKEAVSDIQGEKLNTNDRNQLRPGIGNEPVGGSLGRNTESGPNGGRQGDTLGGTGIRGSNVTQGGNGNT}\\[6pt]
\seqsplit{EKKDGKDNDDKDKKSDDNKGSNGKKVDSLKIPNKITVQSDTIGQSTNGVQNPASSVIGEAQIKQENQINAGEIKAKNGRQGYNTGRGGTVSSGLGGGPGS}\\[6pt]
\seqsplit{KGQSIKDVSDDYKINGLGPIPGKNDDKFQVGNPEGKPLVIAGTKIKGSDGGGHDISSPRQPGGGNVKYRLGSTQPSSHGQHGSRGKSQTGGHGAVNSTSGRGI}\\[6pt]
\seqsplit{QGVKDTMNLVKGGDLNVGVPTGGKSVVKGDQARKNKQTPINVSPSDTGQVRAASAPRTAGNVGRGEGQYKDQRNVIDAGSRGGDGRPGPKGQNGPYAGGD}\\[6pt]
\seqsplit{PHKDSDSRSDVKNVGVEGNSKVVPSGGLNRFKGSSNILTPTNEQIAGIGSGRNAGNKAVEGRTQGSAYSNVGVPGGGRSGTGYSGGGVGRGSYPSQGSRGQSA}
\end{genseq}

\begin{genseq}{Activation steering ($N{=}30$, $\alpha{=}0.00$)}
\seqsplit{HVNNVTRVKFFNSSDMPSAVKYFNKTTHAVLIDIKNQEYNEYTSKDNKVLIKSVGYVKDGVHKYRIILPDISLNSVLESDADRIFDSYLQFFPGGHMDAFTKMS}\\[6pt]
\seqsplit{IANNVQASTSARPGRYWVSELHTDEVPEATPAPTSATIMSKDAHYGILRFEWVSNHDLCESNRIQWKDLVSFPGQHFPLMYFQR}\\[6pt]
\seqsplit{MSTLLRGARQSKVCWAILSRCDQYDYTCQQVKVVTHKLDGKKQIAFHSVGTHPDACDCAIYIPIVPSRSDHGGNAIVV}\\[6pt]
\seqsplit{MATATLLNVESPSLRIPKLKGKRGPYHHRLSYQLYNASQCRVCGEKASGYHFSVSLA}\\[6pt]
\seqsplit{QKTSRMYDRLTSVHLKKMCDSCFGVCPVNAVNVQKCLQTLDLDSIGQPEYTPST}
\end{genseq}

\begin{genseq}{Supervised fine-tuning}
\seqsplit{MADVLDLHEAGGEDFAMDEDGDESIHKLKEKAKKRKGRGFGSEEGSRARMREDYDSVEQDGDEPGPQR}\\[6pt]
\seqsplit{MASCASIDIEDATQHLRDILKLDRPAGGPSAESPRPSSAYNGDLNGLLVPDPLCSGDSTSANKTGLRTMPPINL}\\[6pt]
\seqsplit{APAAAPATAAPAKHAAAPKHRKAQAQMPEAKAAGAKHKKVAKKETAQKAQAAKKHGKKAPAKSTAAKPATKPEAAPAQ}\\[6pt]
\seqsplit{MASCASIDIEDATQHLRDILKLDRPAGGPSAESPRPSSAYNGDLNGLLVPDPLCSGDSTSANKTGLRTMPPINL}\\[6pt]
\seqsplit{PSGTPAPQRYENGAPGGAETGGRFAGANGSVGSWTGQEPAYSGSGSSTGHHTAAPRSSQANLNQSPADYSTVTPQSLQAPTANASN}
\end{genseq}

\begin{genseq}{Natural P-body IDRs}
\seqsplit{VKPESNDKETEAAYESDIPEELCGHHLPQQSLKSYNDSPDVIVEAQFDGSDSEDGHGITQNVLVDGVKKLSVCVSEKGREDGDAPVTKDETTCISQDTRALSEK}\\[6pt]
\seqsplit{LEKLFSGPNAANNKKKKIIPQAKGVVNTAVSAAVQAVRGRGRGTLTRGAFVGATAAPGYIAPGYGTPYGYSTAAPAYGLPKRMVLLPVMKFPTYPVPHYSFF}\\[6pt]
\seqsplit{AGIGPPSEASDSVLMKDAAYPPGPPSNPHVTDTTKKS}\\[6pt]
\seqsplit{MIAPTDSHEEVRSGTSYILPFASRFLSFRADSAHASMVNVPKTRRT}\\[6pt]
\seqsplit{QQIEKEIKEKDSELNQKRPQYIKAKENTSHKIKKLEAAKKSLQNAQKHYKKRKGDMDELEKEMLSVEKARQEFEERMEEESQSQGRDLTLEEN}
\end{genseq}

\newpage
\subsection*{Nuclear pore complex}
\begin{genseq}{RL-SAE, Nuclear pore complex ($N{=}30$)}
\seqsplit{QTLTSSTTAATAFTSTSFSFGLGATTNTGLFGNKTNLGLNTGGGLGVNAGGGGSLGLNTGGGGLFGQKPATAASAGYYQAPTFYAANQQHASQIPDVIRASENF}\\[6pt]
\seqsplit{DAPAPSVNTFTAAPATFGFGNSTFGLSSNNTNTGLFGSKTLGLTNGSGGLGVNTGGGLFGAKPGLGVIGSQPTQTGFHAGAGYGVGAQGVPAGYYHEDEQQAESRKDL}\\[6pt]
\seqsplit{AAASTNSFSTFSFNSTNTFGFGTTNNTGLFGSKTNAGSLGVNTGGLFGNKPAVGVGGGIGVNTGTAGIGVGTPTTSAPAPYYHVDRHREIERVEIEE}\\[6pt]
\seqsplit{PKSSPTNSFNTGVSAFSSFSFGSTNFGQSNTGLFGNKTNLGLNSGTGGLGVNTGGGGLGVNTGGGLFGGKPQVPSAGAPAASASQAGYYHPIHHRVEQIEAE}\\[6pt]
\seqsplit{TEPSNFSLVGTGFSTISSSNFGFGSTNTFGLSNNNTGLFGNTGLNVGGGLFGNKPGGGLGVNNTTGGGYGVNAGQTAGGGFYGMDPASAVLREVIANINNDSIETINQ}
\end{genseq}

\begin{genseq}{RL-GPS, Nuclear pore complex}
\seqsplit{MFSLYTLWSIFFPSFSATWTSIQLSWLNMITSFEFQGIGMSATTASIQLSGVAEFASLSNIFFVAQTMGATTTVEGSQISAGASFQGIESSLNMLAYQNS}\\[6pt]
\seqsplit{TVFLFLGTQANDALGSLTTTLLSYFLFGTPVTLTNFIEYALSLQWTYWAALGPPSSASEQLGHIAFTTNWILLQSLLGAASSAADSVAWFSVNLNALTL}\\[6pt]
\seqsplit{WITAVGSNVGSGTATENWAGWLTSVNGQLSQMGTSSFTTTVTSTGFQPYPFTPFGFTTVENAGSSTATFTQLWWFTGQAELATFASPTAIISYGVSEIQVQ}\\[6pt]
\seqsplit{RTLQTSGTTATAPGWTSWADWTGFIVSGGYSLVTSSFPSSLINGSPTSITAGLSIPSPVWQSLLQSVYSGSPVNVNFQQLPIGTSTDVYFAVYTNDFAQIR}\\[6pt]
\seqsplit{SEIGFTTSSGITNWFLAFFSTSWVTLNSLIGLIFSGSGPSIYPTDSIVGVGNSLVQGLTTATPVPALGNTAFPSVIFGNITSYEGTFQQPAYFMWFMTTA}
\end{genseq}

\begin{genseq}{Activation steering ($N{=}30$, $\alpha{=}0.20$)}
\seqsplit{RSTAAAMATAPPSSQVAPLDFNRDGHVDFLLAHAGGGAAAAIARYTNTGSFNAFTALAGGTFGTVASPVPAFYGGTKFGTLGGAGAAAVPAPIS}\\[6pt]
\seqsplit{MAADNQVLPLVDDDAAQQQGVPLLGAANHSAPQQQSAGFGDLFGAGGAAGGSATAQGSGLGGLLGALTGAPANQGGIGGLLGGGAAMGGTMSGGGLFSGAGGLSLGGGLMA}\\[6pt]
\seqsplit{LSEIASGASRSLAPPVGLTGRARPTNAGSSPERRSAPPSTSVAHRSPSGTMAAASPKSSQSLARTTSTRASFASSSNGSSGPAGGNDK}\\[6pt]
\seqsplit{MKKYKITMNGKSYEVEVAEVSKNATAASAQTAPASSAASQSQVAAPVAAAPAAVAAPAAAPAAVAPAAPAANGSSAGAAIS}\\[6pt]
\seqsplit{LTADLQPGLTGNQSGTASTAAGTAAGTGGTSGSTSTGSTTGTSLTGGSSSLSAAANSSTANGFAGNTASTTNTASTATTGTTTASGTGTASGTATATATATAGAATTLGTGAVGLAGLAAFATALLA}
\end{genseq}

\begin{genseq}{Supervised fine-tuning}
\seqsplit{QNTPRVTTGPTPFSTMPNAAAVAMAATLTQQQQPATGFGTSSGFGCSTTGASTFGFGTTNKPSGWREANDVDFIFFVDLLIIFILYQILYL}\\[6pt]
\seqsplit{MDTEGFGELLQQAEQLAAETEGISELPHVERNLQEIQQAGERLRSRTLTRTSQETADVKASVLLGSRGLDISHISQRLESLSAATTFEPLEPVKD}\\[6pt]
\seqsplit{MAAAEGPVGDGELWQTWLPNHVVFLRLREGLKNQSPTEAEKPASSSLPSSPPPQ}\\[6pt]
\seqsplit{NTPRVTTGPTPFSTMPNAAAVAMAATLTQQQQPATGDKWCNVAELKNRTQLSLFVNKLELSCYCVID}\\[6pt]
\seqsplit{MDTEGFGELLQQAEQLAAETEGISELPHVERNLQEIQQAGERLRSRTLTRTSQETADVKASVLLGSRGLDISHISQRLESLSAATTFEPLEPVKDT}
\end{genseq}

\begin{genseq}{Natural nuclear pore complex IDRs}
\seqsplit{LQDSDEEEEEHPSKTSTKKLKTAPLPPASQTTPLQMALNGKPAPPPQVEKKGQ}\\[6pt]
\seqsplit{MAAAEGPVGDGELWQTWLPNHVVFLRLREGLKNQSPTEAEKPASSSLPSSPPPQ}\\[6pt]
\seqsplit{MYPDLSELLMVSAPNNPTVSGMSDMDYPLQGPGLL}\\[6pt]
\seqsplit{RFPTTLPPPSNVGPILGSPVYSSSPVPSGSPYPNPSFLGTPSHGIQPPAMSTPVCALGNPATQATNMSC}\\[6pt]
\seqsplit{MDTEGFGELLQQAEQLAAETEGISELPHVERNLQEIQQAGERLRSRTLTRTSQETADVKASVLLGSRGLDISHISQRLESLSAATTFEPLEPV}
\end{genseq}

\newpage
\subsection*{Activation domains}
\begin{genseq}{RL-SAE, Activation domains ($N{=}30$)}
\seqsplit{MSDTTNQPDSAPYAADDFMNLDNLDDLFNFDTDMANLTNDDFEDLFFGDMPSDQVPELPLVSPVENEPAETETWGNDWPSDETGS}\\[6pt]
\seqsplit{KPDAPPEIQTTDAMDFNVDFLDISDFNNLSSLSDPDFDIFNSDFFENLTPNTPDIPTPIEDAGRSAETEETWSGNDWGNTE}\\[6pt]
\seqsplit{MPSNPPPTLDDDFNLMGSRDPIDLNELNFDSDLFSDLNFGNLDDIFFDSLLGDKDIPNAAATDQTAPQVSGEEPNGWGGDSWG}\\[6pt]
\seqsplit{GGDAPAPAPVADTMPDDLFLTESDFMGLEFGNLDTDFFTNLAEDDFFTDIEIPNSDIPIADSLPSPASVPQESDDWGSSQWGDT}\\[6pt]
\seqsplit{NQAPVPEPPVLPDTNQAVADDFDTFFSPDLDGFNTDFDLLDTFDNKTPAHQSDDDMNLQESAFAPATDILTDGLNDNSNWGD}
\end{genseq}

\begin{genseq}{Activation steering ($N{=}30$, $\alpha{=}0.43$)}
\seqsplit{VDFDPGGVPDLSQVGSFDDLAGAISSFMDPSFTGTSLDDFDLAGLSDSNPTDLAFPNFVLDSNFASTLPDDITGGYGKFGTANFLGNY}\\[6pt]
\seqsplit{VEPHRSDDPAGQMSLEMMAMLGRMSTLSLDGRAGESTDGMSVENFHATGLADDGDSMEWMFYGLPS}\\[6pt]
\seqsplit{MSGFYNSGYDEIPGLDSTLSAASNTGFGSSSGFLNGGFSEAQNFGGLGNPADLTSGIGGLDSTFSAGTTLVASFGDDDPDFLRISENSLTGTPIVDGNPFLINFVL}\\[6pt]
\seqsplit{MLGLVDDFFFDFDMTEESSLLDGASPDSTASIDLPSNDGTSSPVAVSADPPNQHENLFTNMNETLPNDSTAQFPT}\\[6pt]
\seqsplit{LGPDTDGADDAFTGDEDLVFDAASGLAGDPNIPDVPGLPELFAEDSDEDLDSGGSDPLVPDNGGGFNFDFDPFDDMEPDSFSPDTADFDFTDTGDFDSLGDFETFG}
\end{genseq}

\begin{genseq}{Supervised fine-tuning}
\seqsplit{LEVGSYRMVRAASPDPYSTEETVVVSCCSSVQQQLPAQPLPVPLPCDAATFFEGGNTGSLHRTTGGQY}\\[6pt]
\seqsplit{MTLPGEAGAAGLAPPGHAAAFGGPPGGLLLDALPGPYAAAAAGPLGAAPDRFPADLDLDMFSGSLECDVESIILNDFMDSDEMDFNFDSALPPPPPGLAGAPPPNQSWVP}\\[6pt]
\seqsplit{ALGRNHLADNCYSNYPFPLTSWPCSFSPSQNSSEPFYQQLPLEPPAAKTGCPPLWPNPAGNLYEEKVHVDFNSYVQSPAYHSPQEDPFLFTYASHPHQQY}\\[6pt]
\seqsplit{GLSQSQSPGQTGFLSYGTSFSTPQPGQAPYSYQMQGSSFTTSSGIYTGNNSLTNSSGFNSSQQDYPSYPSFGQGQYAQYYNSSPYPAHYM}\\[6pt]
\seqsplit{VQLIPNFNNWKEPITSDPMESMDTEADSGHLMAMLQEQLDAINMQAGKIQNFLSEIGVAMD}
\end{genseq}

\begin{genseq}{Natural activation domains}
\seqsplit{PEDSLFMFPCGPSGGTSQGYCPGASSGQILMQPPAGNIGTASWSDPCLPELPFPGPFCPQALGHPPGGDGYFPDLFPTPC}\\[6pt]
\seqsplit{GGTPDIPSSGLLSGQAQENPGYPYSDSSSILGENPHIGIDMIDNDQGSSSPSNDEAAMAVIMSLLEADAGLGGPVDFSDLVDFSDLPWPL}\\[6pt]
\seqsplit{MVPSPEAKLTEVDNYHFYSSIPSMEKEVGNCSPHFLNAFEDSFSSILSTEDPNQLTVNSLNSDATVNTDFGDEFYSAFIAEPSISNSMPSPATLSHSLSELLNGPIDVSDLSLCKAFNQNHPESTAEFNDSDSGISLNTS}\\[6pt]
\seqsplit{DLGAHVTKQQSHPEQNSVDYCQQLTVSQGPSPELCDQAIAFSDPLSYFTDLSFSAALKEEQRLDGMLLDDTISPFGTDPLLSATSPAVSKESSRRSSFSSSFSSDDGDEL}\\[6pt]
\seqsplit{LEAGKLPCEDPELDDDFDAHKALEDDEELAHLGLGVDVAKGDDELGTLENLETNDPHLDDLLNGDEFDLLAYTDPELDTGDKKDIFNEHL}
\end{genseq}

\newpage
\subsection*{Repression domains}
\begin{genseq}{RL-SAE, Repression domains ($N{=}30$)}
\seqsplit{MKIDPPPPRVRLPESRALSVPATEGKPQLKVKKEEPMIIQVKKEEPLVIKVEKMEDDDNDDIKSKDEPLQDKNGKTEPMHIKV}\\[6pt]
\seqsplit{LSYDDDRRNTLAPGPTESPSGLRSDAKTEIRTEIKEEPPSIKKEEPLVIKVEKMEEDDDMPDKKNNKEVKDEPLQIKVEEDET}\\[6pt]
\seqsplit{PEKAAEAGPDAVPASPPTAQAPLKGQVKTEVKKEELKIKEEPLVIKVEKMEDDDDMDVKNVKDEPLQIKVEEDSSTNAN}\\[6pt]
\seqsplit{DRQGGPGSAPRPGTGSQTHPTPTDKMTEIKMEPKAEVDVKKEEPLVIKVEKMEDDDDMDVKNDDDMKNKSEPLHIKIEEETAE}\\[6pt]
\seqsplit{GLLLGKTDMEHPTPVQQNGTAREAEEIKGPKTEAKTEVKEEPLVVKVEKMEDDDMDVKNEKNQIKEEPLQIKVEEDTFGV}
\end{genseq}

\begin{genseq}{Activation steering ($N{=}30$, $\alpha{=}0.35$)}
\seqsplit{MLGCLQTGSANHIQWFSTEICPRLCEEEEVNVAECEDTLEYEQIEDDDKIKFIEEIEKLEKEDTEMEKQLEQQEDLEDNEVFVEEEN}\\[6pt]
\seqsplit{MSELTQKEKKPEKKPVEEINVEKEEVFILDEEYEDEEVEEYDLEPSEEIEEFEEIRIREMIEDEIEKEVDEEFSIEEKEEKESSKDE}\\[6pt]
\seqsplit{LKEFEDLHPHQSKQNCMPHFVEEKYEEILNEVKNEIEEPELKVEDYLEDEEESEEVFEEEEFDESKKKNNS}\\[6pt]
\seqsplit{GFIVEENDVDKEPILKEEDWEGVVNPIDEIEEKQVIVEKEVVKEKKVEKKKKVEIKVKEVPVKEEQVKKETDI}\\[6pt]
\seqsplit{MDSNHTDSKPITIEKKEMFSYNDENKTEEEEEEEEEEEEYDDEESIEEFEEKELLSKKESEMQPFHEKIMQEEEELIIEKKEDE}
\end{genseq}

\begin{genseq}{Supervised fine-tuning}
\seqsplit{PGPATLAPSRPLLVPKAERLSPPAPSGSERRLSGDLSSMPGPGTLSVRVSPPQPILSRGRPDSNKTENRRITHISAEQKR}\\[6pt]
\seqsplit{NHWTLGYPNVHEITRSTITEMAAAQGLVDARFPFPALPFTTHLFHPKQGAIAHVLPALHKDRPRFDFANLAVAATQEDPPKMGDLSKLSPGLGSPISGLSKLTPDRKPSR}\\[6pt]
\seqsplit{MDPRGVVPSLWDREHEGKVKNGTCMFVKENNMAEPGSEAPLLSKAGAISQRTEGQREPRDQKTDPQSGAKPLGPPAAASPENTKGKELRILSNFGSTS}\\[6pt]
\seqsplit{MEVMDSCKFSPSELFYDSSCLSSPEGEFPEDFEPRELPVFGAHAPTPPPSTIQR}\\[6pt]
\seqsplit{DDMSDDQGTSVQSKSPDLEASLDTLENNCHVGSDIDFRPKLVNGKGPLDNFLRNRIETSIGQSTVIIDLTEDSNEQPDSL}
\end{genseq}

\begin{genseq}{Natural repression domains}
\seqsplit{QQVSDSTKNGDGTKRPFRQNTHGIQMTSIKKRRSPDDELLYLPVRGRETYEMLLKIKESLELMQYLPQHTIETYRQQQQQQHQHLLQKQTSIQSPSSYGNSSPPLNKMNS}\\[6pt]
\seqsplit{LELTPVQPELQLELVPAAGGGGAAVPGAPAAVVVAPPGYVVVQELMVLPAVAAPAVVAIPGPAGSAALTPARQRRRRRARDRPTICGECGKGFSRSTDLV}\\[6pt]
\seqsplit{CDTVFTHKPLLSSHFDQHLLPQRVSVFKCPSCPLLFAQKRTMLEHLKNTHQSGRLEETAGKGAGGALLTPKTEPEELAVSQGGAAPATEESSSSSEEEEVPSSPEPPRPAKRPRRELGSKGLKGGGGGPGGWTCGLCHSW}\\[6pt]
\seqsplit{NSGREPRTPRTLLSIADILAPRMVPRAPSAPQLPESGPGPTSPLCALEELTSKTFRGLDARALQPSEGRAGPDALGPGPF}\\[6pt]
\seqsplit{HLKKHKAALAAAQYKSSNASETEPNAIKNQTLSASLPLDSTVICSTINKANSVGNGQASQTSQPNYHTKLKKAWLTRHSEEDKNTNKMENSGNSVSEIIKPCSVNLIASTSSDIQNSVDSKIIVDKYVKD}
\end{genseq}